\documentclass[nohyperref]{article}

\usepackage{microtype}
\usepackage{graphicx}
\usepackage{booktabs} 

\usepackage{hyperref}

\usepackage[accepted]{icml2022}

\usepackage{algorithm}
\usepackage{algorithmic}
\usepackage{amsmath}
\usepackage{amssymb}
\usepackage{mathtools}
\usepackage{amsthm}
\usepackage{subdef}
\usepackage{tikz}
\usepackage{multirow}
\usepackage{tabulary}
\usepackage{graphicx}
\usepackage{subcaption}
\usepackage[capitalize,noabbrev]{cleveref}

\theoremstyle{plain}
\theoremstyle{definition}

\theoremstyle{remark}

\newcommand{\figref}[1]{Fig.~\ref{#1}}
\newcommand{\tabref}[1]{Tab.~\ref{#1}}
\newcommand{\secref}[1]{Sec.~\ref{#1}}
\newcommand{\AlgRef}[1]{Algorithm~\ref{#1}}
\newcommand{\equref}[1]{Equ.~(\ref{#1})}

\newcommand{\model}{\mbox{\textsc{Platinum}}}

\usepackage[textsize=tiny]{todonotes}

\definecolor{azure(colorwheel)}{rgb}{0.0, 0.5, 1.0}
\usepackage[font=small]{caption}

\icmltitlerunning{\model: Semi-Supervised Model Agnostic Meta-Learning using Submodular Mutual Information}

\begin{document}

\twocolumn[
\icmltitle{\model: Semi-Supervised Model Agnostic Meta-Learning using Submodular Mutual Information}



\icmlsetsymbol{equal}{*}

\begin{icmlauthorlist}
\icmlauthor{Changbin Li}{equal,yyy}
\icmlauthor{Suraj Kothawade}{equal,yyy}
\icmlauthor{Feng Chen}{yyy}
\icmlauthor{Rishabh Iyer}{yyy}
\end{icmlauthorlist}

\icmlaffiliation{yyy}{University of Texas at Dallas}

\icmlcorrespondingauthor{Suraj Kothawade}{suraj.kothawade@utdallas.edu}
\icmlcorrespondingauthor{Changbin Li}{changbin.li@utdallas.edu}

\icmlkeywords{Machine Learning, ICML}

\vskip 0.3in
]



\printAffiliationsAndNotice{\icmlEqualContribution} 

\begin{abstract}
Few-shot classification (FSC) requires training models using a few  (typically one to five) data points per class. Meta-learning has proven to be able to learn a parametrized model for FSC by training on various other classification tasks. In this work, we propose \model\ (semi-su\textbf{P}ervised mode\textbf{L} \textbf{A}gnostic me\textbf{T}a learn\textbf{I}ng usi\textbf{N}g s\textbf{U}bmodular \textbf{M}utual information
), a novel semi-supervised model agnostic meta learning framework that uses the submodular mutual information (SMI) functions to boost the performance of FSC. \model\ leverages unlabeled data in the inner and outer loop using SMI functions during meta-training and obtains richer meta-learned parameterizations. We study the performance of \model\ in two scenarios - 1) where the unlabeled data points belong to the same set of classes as the labeled set of a certain episode, and 2) where there exist out-of-distribution classes that do not belong to the labeled set. We evaluate our method on various settings on the \textit{mini}ImageNet, \textit{tiered}ImageNet and CIFAR-FS datasets. Our experiments show that \model\ outperforms MAML and semi-supervised approaches like pseduo-labeling for semi-supervised FSC, especially for small ratio of labeled to unlabeled samples.

\end{abstract}

\section{Introduction}

Deep neural networks (DNNs) have proven to be successful in a variety of domains. However, they require large amounts of data, which might not be available for all desired tasks. In such low data regimes, they struggle to perform well. A well known approach to mitigate this problem is meta-learning, which aims to learn from multiple smaller tasks that are related to the target task. The most promising meta-learning techniques that improve the performance of DNNs are gradient based meta-learning methods \cite{finn2017model, rusu2018meta, sun2019meta}. Typically, these methods are designed to operate for few-shot learning. A natural way to improve the performance of meta-learning is by using additional unlabeled data. Semi-supervised techniques are known to use unlabeled data to improve the performance on tasks with relatively small number of labeled data \cite{oliver2018realistic, chapelle2009semi}. 

\begin{figure}[t]
\centering
\begin{subfigure}[]{0.5\textwidth}
\includegraphics[width = \textwidth]{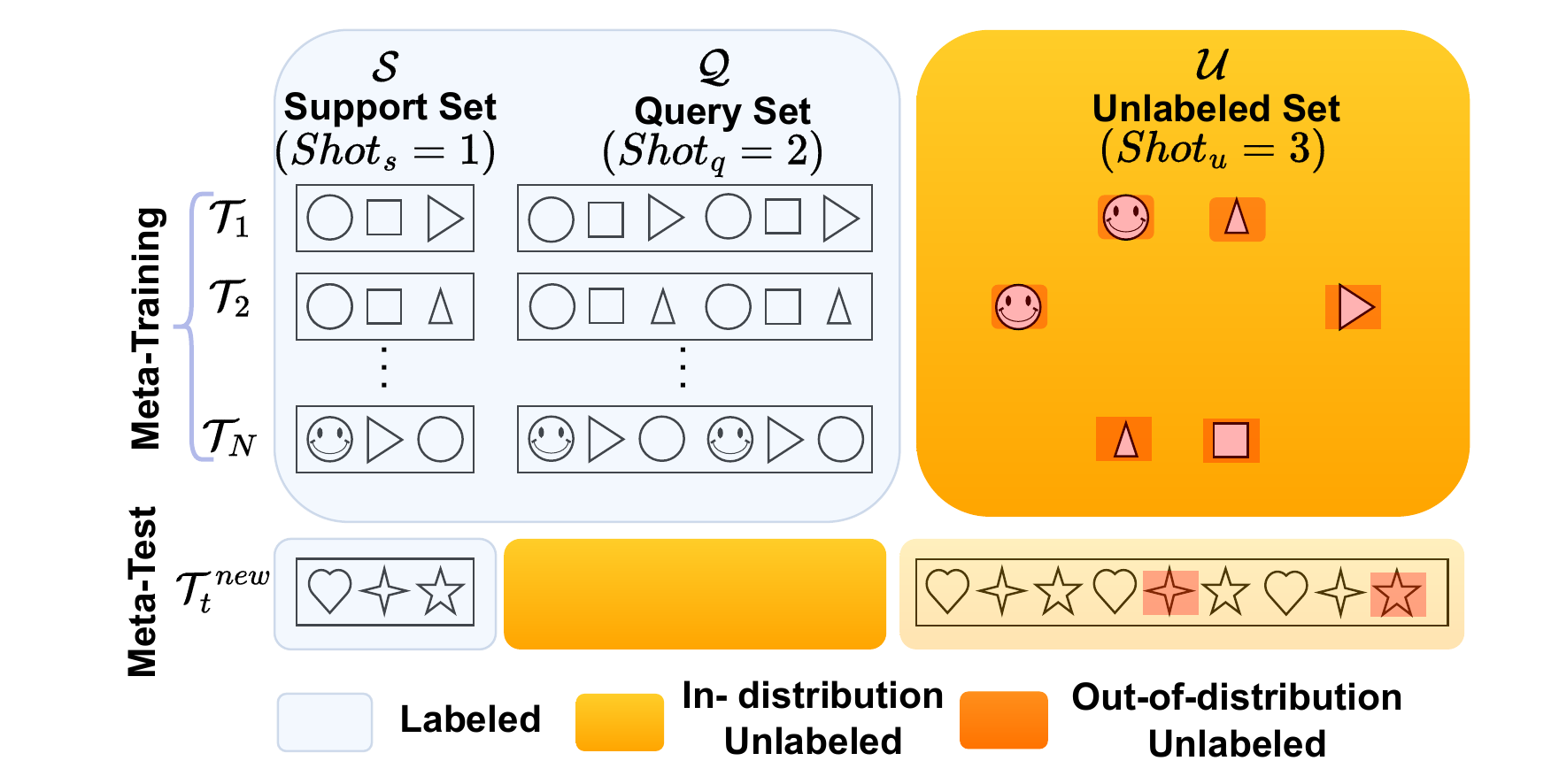}
\end{subfigure} 
\caption{ Semi-supervised few-shot learning setup. During meta-training, the goal is to iterate overs tasks $\Tcal_1 \cdots \Tcal_N$ and meta-learn a parametrization using the support set $\Scal$, query set $\Qcal$, and the unlabeled set $\Ucal$. During meta-testing, the learned parametrization is used as an initialization and a model is trained using the $\Scal$ and $\Ucal$ to perform well on $\Qcal$. In any task, $\Ucal$ may contain data points that are out-of-distribution, \ie not pertinent to the classes of data points in $\Scal$. 
\vspace{-3ex}}
\label{fig:Intro}
\end{figure}

\begin{figure*}[!htbp]
    \centering
    \includegraphics[width = \textwidth]{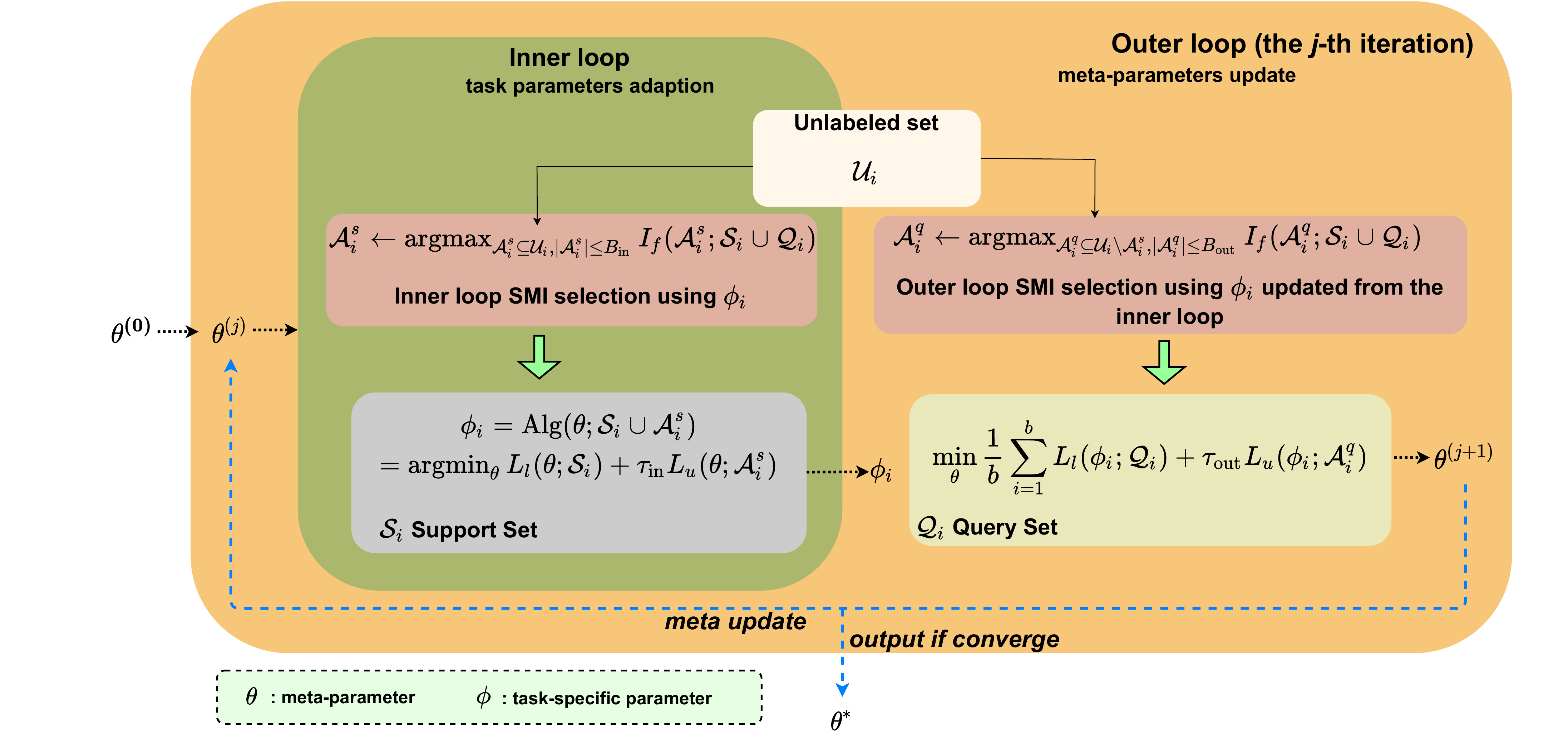}
    \caption{Overview of our \model{} framework that solves a semi-supervised few-shot learning problem. For a specific task $\mathcal{T}_i$, in each inner loop and outer loop gradient update step, we select a subset from the unlabeled set by maximizing the per-class SMI function (see \AlgRef{alg-ssl-fsl}). In each \textit{inner loop} step, the selected subset $\mathcal{A}^{s}_i$ and support set $\Scal_i$ will be used to update model parameters $\phi_i$ using \equref{param-adaptation-ssl}. In the \textit{outer-loop} of the meta-training stage, another subset $\mathcal{A}^{q}_{i}$ will be selected after inner loop selection according to the updated model parameters $\phi_i$. Meta-parameters $\theta$ would be updated based on $\mathcal{A}^{q}_{i}$ and the query set $\Qcal_i$ using  \equref{meta-objective-ssl}. It should be noted that, as temperature annealing coefficients, $\tau_{\text{in}}$ is a function of inner step $t_{\text{in}}$, and $\tau_{\text{out}}$ is a function of the global iteration index.} 
    \label{fig:overview}
    \vspace{-4mm}
\end{figure*}

In this paper, we focus on few-shot classification using Model Agnostic Meta-Learning (MAML) and improve it via semi-supervised learning (see \figref{fig:Intro}). In many realistic classification tasks, although the labeled data is scarce, there is plenty of unlabeled data available for training a classifier. Towards this goal, we propose \model, a novel framework that embeds semi-supervision in the MAML framework by using \textbf{submodular mutual information (SMI)} \cite{smi} functions as per-class acquisition functions. We observe that embedding semi-supervision in the MAML framework is non-trivial, since simply using a pseudo-labeling approach in the inner loop does not improve the performance. This lack of improvement occurs due to either noisy pseudo-labels or class imbalance caused due to pseudo-labels being confident only for certain classes. To overcome these issues, \model\ uses a class-wise unique instantiations of SMI functions to provide per-class semi-supervision. Furthermore, these per-class acquired subsets are diverse, leading to a richer meta-learned parameterization (see Fig.~\ref{fig:overview}).  

\subsection{Related Work}
\textbf{Few-shot learning.} There has been an extensive amount of work in few-shot learning, which has mainly revolved around supervised learning. Although our framework is embedding semi-supervision into MAML which belongs to the gradient descent family of methods, the few-shot learning literature can be broadly divided into the following categories: 1) Metric learning methods \cite{vinyals2016matching, snell2017prototypical} which deal with learning a similarity space where the task can be efficiently done with a few labeled data points. 2) Memory networks \cite{munkhdalai2017meta, santoro2016meta, oreshkin2018tadam, mishra2017simple}, which focus on learning to store ``experience" from previously observed tasks in the interest of generalizing to newer tasks. 3) Gradient based meta-learning methods \cite{finn2017model, finn2018probabilistic, antoniou2018train, ravi2016optimization, grant2018recasting, sun2019meta, killamsetty2020nested, zhao2022adaptive} which aim to meta-learn a model in the outer loop that is used as a starting point in the inner loop for a new few-shot task. In addition, some researches use pre-training strategy to boost the performance~\cite{chen2019closer, tian2020rethinking, wang2021role}. The \model\ framework embeds semi-supervision for gradient descent based methods that use an inner-outer bi-level optimization.\\

\textbf{Semi-supervised learning (SSL).} The goal of SSL methods is to leverage unlabeled data alongside the labeled data to obtain a better representation of the dataset than supervised learning \cite{oliver2018realistic}. The most basic SSL method, Pseudo-labeling \cite{lee2013pseudo} uses model predictions as target labels as a regularizer, and a standard supervised loss function for the unlabeled dataset. Some SSL methods like $\Pi$-Model \cite{laine2016temporal, sajjadi2016regularization} and Mean Teacher \cite{tarvainen2017mean} use consistency regularization, by using data augmentation and dropout techniques. Mean Teacher obtains a more stable target output by using an exponential moving average of parameters across previous epochs. Virtual Adversarial Training (VAT) \cite{miyato2018virtual} uses an effective regularization technique that uses slight perturbations such that the prediction of the unlabeled samples is affected the most. More recent techniques like FixMatch \cite{sohn2020fixmatch}, MixMatch \cite{berthelot2019mixmatch} and UDA \cite{xie2019unsupervised} use data augmentations like flip, rotation, and crops to predict pseudo-labels. In this paper, we propose a new SSL technique that uses class-wise instantiations of SMI functions that mitigates the issue of class-imbalance in selected subsets and is comparatively robust to OOD classes in the unlabeled set.

\textbf{Semi-supervised few-shot learning.}
There are two categories for semi-supervised few-shot learning: 1) \textit{meta-learning based}: \citet{ren2018meta} propose new extensions of Prototypical Networks \cite{snell2017prototypical} by viewing each prototype as a cluster center and tuning the cluster locations using soft K-means such that the data points in support and unlabeled sets fit better. \citet{liu2018learning} learn a graph construction module to propagate labels from labeled examples to unlabeled examples. In addition, \citet{li2019learning} propose learning to self-train (LST) which aims to meta-learn how to cherry-pick and label data points from the unlabeled set and optimizes weights of these pseudo-labels. However, their method is on the top of a pretrained meta-transfer learning \cite{sun2019meta} model which requires the labels across all training tasks to be known beforehand. Unfortunately, such meta-data about the dataset may not be available in most realistic scenarios. On the other hand, our \model\ framework does not require a pre-trained network or any meta-data for embedding semi-supervision in gradient descent based methods. 2) \textit{Transfer learning based} \cite{yu2020transmatch,Wang_2020_CVPR,Huang_2021_ICCV, huang2021ptn,lazarou2021iterative}: this is the main focus in more recent works. Similar to LST~\cite{li2019learning}, they assume all examples of base classes are labeled so that a feature extractor could be pretrained based on them. In contrast, we assume there are few examples per class are labeled during both meta-training and meta-test, which is more realistic than that in transfer learning based approaches. In addition, transfer learning based approaches do not leverage episodes training strategy, which is different from ours. 

\textbf{Data subset selection (DSS).} DSS using submodular functions has been studied in the context of various applications like video summarization \cite{kaushal2020realistic, kaushal2019framework}, image-collection summarization \cite{tschiatschek2014learning, kothawade2020deep}, efficient learning \cite{kaushal2019learning,killamsetty2020glister,killamsetty2021grad,liu2017svitchboard}, targeted learning \cite{kothawade2021prism, kothawade2021talisman}, \etc\  Recently, \cite{kothawade2021prism} used the SMI functions for improving the performance of rare classes in the context of image classification, and \cite{kothawade2021talisman} used them for mining rare objects and slices for improving object detectors. \cite{kothawade2021similar} used the submodular information measures as acquisition functions for active learning in scenarios with class imbalance, redundancy and OOD data. \cite{killamsetty2020glister,killamsetty2021grad} studied the role of submodular functions and coresets for compute-efficient training of deep models.

\subsection{Our Contributions}
The following are our main contributions: 1) Given the limitations of existing approaches, we propose \model\ (see \secref{sec:method}) that uses per-class semi-supervision using SMI functions, thereby preventing class-imbalance in the selected subset for semi-supervision. 2) Our framework learns richer parameterizations by embedding semi-supervision in the inner and outer loop of MAML. 3) We conduct extensive experiments on  \textit{mini}ImageNet \cite{vinyals2016matching},  \textit{tiered}ImageNet \cite{ren2018meta}, and CIFAR-FS \cite{bertinetto2018meta} datasets where the unlabeled set has in-distribution and out-of-distribution (OOD) classes. 4) We conduct various ablation experiments where we study the effect of varying the: i) ratio of labeled and unlabeled data points, ii) number of OOD classes, and iii) inner and outer loop selection for semi-supervision.

\section{Preliminaries} \label{sec:preliminaries}

\subsection{Model Agnostic Meta Learning (MAML)}
MAML~\cite{finn2017model} is a representative of gradient-based meta-learning approach, its goal is to obtain optimal initial model parameters for \textit{unseen} tasks. Suppose there are a set of meta-training tasks sampled from a task distribution $p(\mathcal{T})$. Each task $\mathcal{T}_i$ is split into support (training) set and query (validation) set $\{\mathcal{S}_i, \mathcal{Q}_i\}$. As a bi-level optimization problem, in the \textit{inner loop}, MAML adapts the task-specific model parameters $\phi_i$ from initialization parameters $\theta$ for task $\mathcal{T}_i$ based on its support set: $\phi_i = \argmin_{\theta} \left[L({\theta}; \mathcal{S}_i)\right]$ ($L$ is the loss of model parameterized by $\theta$ on data $\mathcal{S}_i$). The loss of adapted model $\phi_i$ on the corresponding query set $L(\phi_i; \mathcal{Q}_i)$ is used to evaluate the performance. 
 In the \textit{outer loop}, the averaged query set loss is minimized to obtain the optimal initial parameters.
 Therefore, the objective function could be formulated as follows:
\begin{equation}
\label{meta-objective}
\theta^*= \argmin_{\theta \in {\Theta}}{\mathbb{E}_{\mathcal{T}_i\sim p(\mathcal{T})}}\left[L(\text{Alg}(\theta; \mathcal{S}_{i}); \mathcal{Q}_i)\right]
\end{equation}
where $\text{Alg}(\theta; \mathcal{S}_{i})$ corresponds to single or multiple gradient descent steps in the inner-level task-specific adaptation. In case of single-step gradient update, $\text{Alg}(\theta; \mathcal{S}_{i})$ can be specified as following:
\begin{equation}
\label{param-adaptation}  
\phi_i = \text{Alg}(\theta; \mathcal{S}_{i}) \approx \theta - \alpha \nabla_{\theta}{L}(\theta; \mathcal{S}_{i})
\end{equation}
where $\alpha $ is a learning rate. The learned meta-parameters $\theta^{*}$ from outer-level will be leveraged as model initialization for the \textit{unseen} tasks during meta-test stage. A table of notations with corresponding explanations is given in Appendix~\ref{app:notations}.

\subsection{Submodular Mutual Information} 
\textbf{Submodular functions.} Submodular functions \cite{tohidi2020submodularity, bach2011learning, bach2019submodular} have been widely used for data subset selection as they naturally model properties like coverage, representation, diversity, \etc. Given a ground-set of $n$ data points $\Vcal = \{1,2,3, \cdots, n\}$, and a set function $f: 2^{\Vcal} \xrightarrow{} \mathbb{R}$. The set function $f$ is known to be submodular \cite{fujishige2005submodular} if for $x \in \Vcal$, $f(\Acal \cup x) - f(\Acal )\geq f(\Bcal \cup x) - f(\Bcal)$, $\forall \Acal \subseteq \Bcal \subseteq \Vcal$ and $x \notin \Bcal$. We use two well known submodular functions in this work, facility location (\textsc{Fl}) and graph-cut (\textsc{Gc}) (see \tabref{tab:submod_inst}(a)) that can be instantiated using a similarity kernel containing pairwise similarities between all data points. In general, submodular functions admit a $1-\frac{1}{e}$ constant factor approximation \cite{nemhauser1978analysis} for cardinality constraint maximization. Furthermore, they can be optimized in \emph{near-linear} time using greedy algorithms \cite{mirzasoleiman2015lazier}.
\\

\textbf{Submodular Mutual Information (SMI).}
In this work, we use the SMI instantiations of the above submodular functions to provide semi-supervision. Particularly, we use \textsc{Flmi} and \textsc{Gcmi} where the underlying submodular function is \textsc{Fl} and \textsc{Gc} respectively (see \tabref{tab:submod_inst}(b)). The SMI functions can be used to select data points that are semantically \emph{similar} to the data points in a given query set \cite{kothawade2021similar, kothawade2021prism, kothawade2021talisman}. To obtain pseudo-labels, we use exemplars from a particular class in the query set used to instantiate an SMI function. The subset obtained by optimizing this SMI function is then assigned labels of the class of the data points used in the query set. Formally, the submodular mutual information (SMI) is defined as $I_f(\Acal; \Rcal) = f(\Acal) + f(\Rcal) - f(\Acal \cup \Rcal)$, where $\Rcal$ is a query set. Note that \cite{smi, kothawade2021prism} propose a few other SMI functions. However, we use only the \textsc{Flmi} and \textsc{Gcmi} variants in the interest of scalability (see \secref{sec:smi_scalability}). 

\begin{table}[!htb]
    \caption{Instantiations of different submodular functions. 
    }
    \label{tab:submod_inst}
    \begin{subtable}{.40\linewidth}
      \centering
        \caption{Instantiations of submodular functions.}
        \begin{tabular}{|l|l|}
        \hline
        \textbf{SF} & \textbf{$f(\Acal)$} \\ \hline
        \scriptsize{\textsc{Fl}}       & \scriptsize{$\sum\limits_{i \in \Ucal} \max\limits_{j \in \Acal} S_{ij}$}   \\
        \scriptsize{\textsc{Gc}}       & \scriptsize{$\sum \limits_{i \in \Acal, j \in \Ucal} S_{ij} -$} \\ & \scriptsize{$ \sum\limits_{i, j \in \Acal} S_{ij}$} \\
        \hline
        \end{tabular}
        \vspace{0.5ex}
    \end{subtable} 
    \begin{subtable}{.40\linewidth}
      \centering
        \caption{Instantiations of SMI functions.}
        \begin{tabular}{|c|c|c|}
        \hline
        \textbf{SMI} & \textbf{$I_f(\Acal;\Rcal)$} \\ \hline
        \scriptsize{FLMI}             & \scriptsize{$\sum\limits_{i \in \Rcal} \max\limits_{j \in \Acal} S_{ij} + $ $ \sum\limits_{i \in \Acal} \max\limits_{j \in \Rcal} S_{ij}$}                \\
        \scriptsize{GCMI}              & \scriptsize{$2 \sum\limits_{i \in \Acal} \sum\limits_{j \in \Rcal} S_{ij}$}                \\
         \hline
        \end{tabular}
    \end{subtable}%
\end{table}

\section{\model: Our Semi-Supervised Meta-Learning Framework} \label{sec:method}


In this section, we define the semi-supervised few-shot classification setting considered in this work (see \figref{fig:Intro}). 
We start with $N$ meta-training tasks $\Tcal_1 \cdots \Tcal_N $. For each task $\Tcal_i$, we have, $\{(\mathcal{S}_i, \mathcal{Q}_i, \mathcal{U}_i)\}_{i=1}^{N}$, where $\Scal$ is the labeled support set, $\Qcal$ is the query set with unseen data points for test, and $\Ucal$ is the unlabeled set. In our experiments (\secref{sec:experiments}), similar to \cite{ren2018meta}, we consider both settings, where $\Ucal$ \emph{does} or \emph{does not} consist of OOD classes. 

The goal of our method is to obtain the optimal initial parameters that result into faster adaptation of the classifier to a new task. To do so, we minimize the following meta-training objective:

\begin{equation}
\begin{aligned}
\label{meta-objective-ssl}
     \theta^*= \argmin_{\theta \in \boldsymbol{\Theta}}{\mathbb{E}_{\mathcal{T}_i \sim p(\mathcal{T})}\mathcal{J}(\theta)}\text{\hspace{1.7cm}}\\
    \text{where }\mathcal{J}(\theta) = L({\color{azure(colorwheel)}\text{Alg}(\theta; \mathcal{S}_{i}\cup\mathcal{A}^{s}_{i}); \mathcal{Q}_{i}\cup\mathcal{A}^{q}_i})
\end{aligned}
\end{equation}

Here, $\mathcal{A}^{s}_{i} \subseteq \Ucal_i$ and  $\mathcal{A}^{q}_{i} \subseteq \Ucal_i$ are selected subsets with hypothesized labels in the \textit{inner loop} and \textit{outer loop}, respectively (see \secref{sec:meta_training}).  $\color{azure(colorwheel)}\text{Alg}(\theta; \mathcal{S}_{i}\cup\mathcal{A}^{s}_{i})$ corresponds to single or multiple updates on support set $\Scal_i$, and hypothesized labeled subset $\Acal_i^s$ for task $\mathcal{T}_i$ in the \textit{inner loop}. We consider multiple steps in the inner loop in practice. 

\begin{equation}
    \label{param-adaptation-ssl}
    \begin{aligned}
    \phi_i &= {\color{azure(colorwheel)}\text{Alg}(\theta; \mathcal{S}_{i}\cup\mathcal{A}^{s}_{i})}\\
    &= \operatorname{argmin}_{\theta} L\left(\theta ; \mathcal{S}_{i}\cup\mathcal{A}^{s}_{i}\right)\\
    &= \phi_i - \nabla_{\phi}  L\left(\theta ; \mathcal{S}_{i}\cup\mathcal{A}^{s}_{i}\right)
    \end{aligned}
\end{equation}


In addition to the inner loop selection, \model\ embeds semi-supervision for the outer-loop selection. We do so since the outer-level also corresponds to the meta-training objective of generalizing well, especially when data is scarce to update meta-parameters. The model parameters updated from inner loop $\phi_i$ would be used to do outer loop selection. We perform the outer-loop update as follows:
\begin{equation}
\label{eq:outer_param_update}
\begin{aligned}
{\mathcal{J}(\theta)} &= L({\phi_i}; (\mathcal{Q}_{i} \cup \mathcal{A}^{q}_{i})) \\
\end{aligned}
\end{equation}
For meta-testing, we sample a new \emph{unseen} task $\Tcal^{new}$. The \emph{unseen} task for meta-testing is made of disjoint set of data points and classes from the tasks seen during meta-training. Next, we use the parameters from obtained from the meta-training stage and initialize a model and train it on $\{\Scal^{new}, \Ucal^{new}\}$. Finally, we evaluate the model on $\Qcal^{new}$, and report the average accuracy across all unseen tasks.

\begin{algorithm}[t]
\small
\caption{\model  \text{ }(Meta-Training)}
\begin{algorithmic}[1]
\label{alg-ssl-fsl}
\REQUIRE task distribution: $p(\mathcal{T})$, Base model with parameters $\theta$, Batch size of tasks: $b$, Budge of selected samples from unlabeled set: $B_{\text{in}}$, $B_{\text{out}}$

\STATE Randomly initialize $\theta$
\WHILE{not converge}
\STATE Sample a batch of tasks $\{\mathcal{T}_i\}_{i=1}^{b} \sim p(\mathcal{T})$
    \FOR{each task $\mathcal{T}_i=\{\mathcal{S}_i,\mathcal{Q}_i, \mathcal{U}_i\}, i\in [b]$ }
    
    
    \STATE Initialize model parameters $\phi_i \leftarrow \theta$
    
        \FOR{each inner step $t$}
        \STATE $\Pcal_{\Ucal_i} \leftarrow \phi_i(\Ucal_i)$
        \STATE $\Xcal \leftarrow$ \textsc{Cosine\_Similarity}($\Pcal_{\Ucal_i}, \{\Pcal_{\Scal_i} \cup \Pcal_{\Qcal_i}\}$)
        \STATE Instantiate a submodular function $f$ based on $\Xcal$.
        
        {\color{azure(colorwheel)} $/*$ \textit{inner loop selection} $*/$}
        
        \STATE $\mathcal{A}^{s}_{it} \leftarrow \operatorname{argmax}_{\mathcal{A}_{it}^s \subseteq \mathcal{U}_i,|\mathcal{A}^s_{it}| \leq B_{\text{in}}} I_{f}(\mathcal{A}^s_{it}; \mathcal{S}_i \cup \mathcal{Q}_i)$ \textcolor{azure(colorwheel)}{\{Acquire subset with hypothesized labels using per-class greedy maximization, Equ. \eqref{eq:per_class_smi}\}}
        
        \STATE $\phi_i \leftarrow \phi_i - \nabla_{\phi}  L\left(\theta ; \mathcal{S}_{i}\cup\mathcal{A}^{s}_{i}\right)$
        \textcolor{azure(colorwheel)}{Update $\phi_i$ by gradient descent, Equ. (\ref{param-adaptation-ssl})}
        \STATE $\Acal_i^s \leftarrow \Acal_i^s \cup \Acal_{it}^s$
        \ENDFOR
    \STATE $\Pcal_{\Ucal_i \backslash \Acal_i^s} \leftarrow \phi_i(\Ucal_i \backslash \Acal_i^s)$ 
    \STATE $\Xcal \leftarrow$ \textsc{Cosine\_Similarity}($\Pcal_{\Ucal_i \backslash \Acal_i^s}, \{\Pcal_{\Scal_i} \cup \Pcal_{\Qcal_i}\}$)
    
    {\color{azure(colorwheel)} $/*$\textit{outer loop selection}$*/$}
    
    \STATE $\mathcal{A}^{q}_{i} \leftarrow \argmax_{\mathcal{A}^{q}_{i}\subseteq \mathcal{U}_i\backslash\mathcal{A}^s_i, |\mathcal{A}^{q}_{i}|\leq B_{\text{out}}}I_{f}(\mathcal{A}^{q}_{i}; \mathcal{S}_i\cup \mathcal{Q}_i)$ \textcolor{azure(colorwheel)}{\{Acquire subset with hypothesized labels using per-class greedy maximization, Equ. \eqref{eq:per_class_smi}\}}
    \ENDFOR
    
$/*$\textit{meta update (outer loop)} $*/$

\STATE Obtain $\theta^{(t+1)}$ by Equ. \eqref{meta-objective-ssl} using $\{\mathcal{Q}_i\cup \mathcal{A}^{q}_i\}_{i=1}^{b}$
\ENDWHILE
\STATE Return the meta-learned parameters $\theta$.
\end{algorithmic}
\end{algorithm}

\subsection{Leveraging Full Potential of SSL during Meta-Training} \label{sec:meta_training}

In this section, we discuss the meta-training pipeline of \model. Particularly, we emphasis on \emph{inner loop} and \emph{outer loop} semi-supervision embeded using \emph{class-wise} SMI instantiations. We detail our pipeline in \AlgRef{alg-ssl-fsl}.

For any task $\Tcal_i  \sim p(\mathcal{T})$, we first initialize the model parameters $\phi_i \leftarrow \theta$, where $\theta$ is meta-learned by optimizing the outer loop on the previous tasks. Using parameters $\phi_i$, we compute an embedding containing class probabilities for each data point belonging to the unlabeled set $\Ucal_i$. Emprically, we found out that using the class probabilities based on the classes belonging to $\Tcal_i$ was as effective as using a larger and computationally expensive embedding like last layer features or gradients. Since the support and query set have labels during meta-training, we use a one-hot vector to represent the data points in $\Scal_i$ and $\Ucal_i$, where the probabilty of the class corresponding to the label is set to one. Next, we compute a pairwise similarities $\Xcal_{pq}$, where $p \in \{\Scal_i \cup \Qcal_i\}, q \in \Ucal_i$. For each class $c$, we instantiate an SMI function $I_f^c$ (\tabref{tab:submod_inst}) by using a sub-matrix $\Xcal^c$ with pairwise similarities $\Xcal^c_{pq}$ such that $p$ belongs to class $c$, $\forall p \in \{\Scal_i \cup \Qcal_i\}$. We then maximize $I_f^c$ with a budget of $B/C$ as follows:

\begin{equation}
    \label{eq:per_class_smi}
    \mathcal{A}^c \leftarrow \argmax_{\mathcal{A}^c\subseteq \mathcal{U}_i, |\mathcal{A}^c|\leq B/C}I_{f}^c(\mathcal{A}^c; \mathcal{S}_i\cup \mathcal{Q}_i) 
\end{equation}

where, $C$ is the number of classes in the current task $\Tcal$. Since $\Acal^c$ is obtained by optimizing the SMI function $I_f^c(\Acal^c;\Rcal)$, where $\Rcal$ contains data points \emph{only} from class $c$, we assigned the hypothesized label $c$ to all data points in $\Acal^c$. We obtain $\Acal^s = \{\Acal^1 \cup \Acal^2 \cdots \cup \Acal^C\}$ by repeating the selection for each class. Finally, we update $\phi_i$ in the inner loop by using gradient descent on $\Scal_i$ and $\Acal^s_i$. Note that $\phi_i$ is updated for $T_{\text{in}}$ steps in the inner loop. Similarly, we obtain $\Acal^q_i$ by using the SMI functions to embed semi-supervision in the outer loop and update $\theta$ by gradient descent on $\Qcal_i$ and $\Acal^q_i$. We summarize the inner loop and outer loop optimization problems in \equref{eq:inner_sel} and \equref{eq:outer_sel} respectively, and discuss them in more detail:

\begin{equation}
    \mathcal{A}^{s}_{i} \leftarrow \argmax_{\mathcal{A}^{s}_{i}\subseteq \mathcal{U}_i, |\mathcal{A}^{s}_{i}|\leq B_{\text{in}}}I_{f}(\mathcal{A}^{s}_{i}; \mathcal{S}_i\cup \mathcal{Q}_i) 
    \label{eq:inner_sel}
\end{equation}

\begin{equation}
    \mathcal{A}^{q}_{i} \leftarrow \argmax_{\mathcal{A}^{q}_{i}\subseteq \mathcal{U}_i\backslash\mathcal{A}^s_i, |\mathcal{A}^{q}_{i}|\leq B_{\text{out}}}I_{f}(\mathcal{A}^{q}_{i}; \mathcal{S}_i\cup \mathcal{Q}_i)
    \label{eq:outer_sel}
\end{equation}

where $B_{\text{in}}$ and $B_{\text{out}}$ are selection budget per class.

\textbf{Inner loop.}
Although MAML could achieve single step gradient update in the inner loop, it is not common to have good adaptation in practice especially considering the involving of additional unlabeled set. To illustrate this clearly, we assume there are $T_{\text{in}}$ steps during the model adaptation in the inner loop. Inspired from ~\cite{lee2013pseudo}, we add some unlabeled examples to update a task-specific model $\phi_i$. Different from ~\cite{lee2013pseudo}, we do not use all examples in the unlabeled set because efficiency matters in meta-learning training procedure. Therefore, the loss function in the inner loop is formulated as below:
\begin{equation}
    L\left(\theta ; \mathcal{S}_{i}\cup\mathcal{A}^{s}_{i}\right) = L_{l}\left(\theta ; \mathcal{S}_{i}\right)+\tau_{\text{in}} L_{u}\left(\theta ; \mathcal{A}^{s}_{i}\right)
\end{equation}
where $\mathcal{A}^{s}_{i}$ is the selected examples with pseudo labels.We define $L_{l}$ as the loss based on examples with true labels, and $L_{u}$ as the loss function based on hypothesized labeled examples. Similar formulations have been used in conventional semi-supervised learning approaches, such as Pseudo-Label~\cite{lee2013pseudo} and VAT~\cite{miyato2018virtual}. $\tau_{\text{in}}$ is a temperature annealing coefficient: 
\begin{equation}
\small 
\tau_{\text{in}}(t)= \begin{cases}0 & t<2 \\ \exp{(-5(1 - \frac{t}{T_{\text{in}}})^2)} & 2 < t\leq T_{\text{in}}\end{cases}
\end{equation}
Note that we consider multi inner step updates, and SMI subset selection happens in each step  (line 6-13 in Algorithm~\ref{alg-ssl-fsl}).


\textbf{Outer loop.}
Considering the meta-parameters are updated in the outer loop based on the labeled query set, and there are few labeled examples per class, it is beneficial to augment the query set aiming to generalize well for novel class in the meta-test stage. Considering the unlabeled examples, the loss function in the outer loop could be:  
\begin{equation}
\begin{aligned}
{\mathcal{J}(\theta)} &= {L}({\phi_i}; \mathcal{Q}_{i}\cup  \mathcal{A}^{q}_{i}) \\
&=L_{l}\left(\phi_{i} ; \mathcal{Q}_{i}\right)+ \tau_{\text{out}} L_{u}\left(\phi_{i} ; \mathcal{A}^{q}_{i}\right) \\
\end{aligned}
\end{equation}
where $\mathcal{A}^{q}_{i}$ is the selected examples with pseudo labels, and $\tau_{\text{out}}$ is a temperature annealing coefficient: 
\begin{equation}
\small 
\tau_{\text{out}}(j)= \begin{cases}\exp{(-5(1 - \frac{t}{T_{\text{warm}}})^2)} & 0 < j\leq T_{\text{warm}}  \\ 1 & T_{\text{warm}}<j\leq T_{\text{out}} \end{cases}
\end{equation}

$T_{\text{out}}$ is the total number of epochs during meta-training procedure. $T_{\text{warm}}$ is a warm starting epoch index. More detailed selection process explanation is given in Appendix~\ref{app:inner_outer}.

\begin{table}[!htbp]
\caption{Running time (training time of 100 tasks) comparison on \textit{mini}ImageNet domains for 1-shot (5-shot) 5-way experiments without OOD classes in unlabeled set.}
\label{tab: running_time}
\begin{center}
\begin{small}
\begin{sc}
\begin{tabular}{lcc}
\toprule
Methods & 1-shot (s) & 5-shot (s)  \\
\midrule
MAML             & 23.92  & 49.83\\
\textsc{Gcmi}    & 27.33  & 58.91\\
\textsc{Flmi}    & 28.94  & 56.04\\ 
\bottomrule
\end{tabular}
\end{sc}
\end{small}
\end{center}
\vspace{-5mm}
\end{table}

\begin{table*}[!htbp]
\caption{Few-shot classification accuracy ($\%$) on \textit{mini}ImageNet. ({\textdagger}: only supervised setting is considered.)} 
\label{tab:classification_main_1}
    \small
    \centering
    \begin{small} 
    \begin{tabular}{l|cc|cc}
        \toprule
        & \multicolumn{2}{c|}{\textit{1-shot}} & \multicolumn{2}{c}{\textit{5-shot}}\\
        Methods & \textit{w/o OOD} & \textit{w/ OOD} & \textit{w/o OOD} & \textit{w/ OOD}\\
        \midrule
    
        Soft k-Means~\cite{ren2018meta}         & 24.61{\tiny $\pm$0.64} &	23.57{\tiny $\pm$0.63} &	38.20{\tiny $\pm$1.64} &	38.07{\tiny $\pm$1.53} \\
        Soft k-Means+Cluster~\cite{ren2018meta} & 15.76{\tiny $\pm$0.59} &	9.77{\tiny $\pm$0.51}  &	33.65{\tiny $\pm$1.53} &	30.47{\tiny $\pm$1.42} \\
        Masked Soft k-Means~\cite{ren2018meta}  & 25.48{\tiny $\pm$0.67} &	25.03{\tiny $\pm$0.68} &	39.33{\tiny $\pm$1.55} &	38.48{\tiny $\pm$1.74} \\
        TPN-semi~\cite{liu2018learning} &  40.25{\tiny $\pm$0.92}  &	26.70{\tiny $\pm$0.98} &	46.27{\tiny $\pm$1.67} &	36.81{\tiny $\pm$0.87} \\
        LST(\textit{small})~\cite{li2019learning}   & 37.65{\tiny $\pm$0.78}  & 37.82{\tiny $\pm$0.91}  &  61.50{\tiny $\pm$0.92}  & 57.67{\tiny $\pm$0.85}  \\
        LST(\textit{large})~\cite{li2019learning}   & 41.36{\tiny $\pm$0.98}  & 39.32{\tiny $\pm$0.95}  &  61.51{\tiny $\pm$0.98}  & 59.24{\tiny $\pm$0.95}  \\
        {MAML}{}\textsuperscript{\textdagger} \cite{finn2017model}     &   35.26{\tiny $\pm$0.85} &	35.26{\tiny $\pm$0.85} &	60.22{\tiny $\pm$0.83} &	60.20{\tiny $\pm$0.83} \\
        VAT \cite{miyato2018virtual}     &  36.55{\tiny $\pm$0.86} &	34.03{\tiny $\pm$0.84} &	61.60{\tiny $\pm$0.83} &	61.24{\tiny $\pm$0.88}  \\
        PL \cite{lee2013pseudo}      &  37.71{\tiny $\pm$0.94} &	35.16{\tiny $\pm$0.85} &	60.64{\tiny $\pm$0.92} &	60.31{\tiny $\pm$0.87} \\
        \hline 
        \textsc{Gcmi} (ours)  &  41.94{\tiny $\pm$0.96} &	\textbf{42.57}{\tiny $\pm$0.93} &	63.62{\tiny $\pm$0.95} &	\textbf{63.54}{\tiny $\pm$0.94} \\
        \textsc{Flmi} (ours) &  \textbf{42.27}{\tiny $\pm$0.95} &	41.53{\tiny $\pm$0.97} &	\textbf{63.80}{\tiny $\pm$0.92} &	63.44{\tiny $\pm$0.99} \\
        \bottomrule
    \end{tabular}
    \end{small}
\end{table*}

\begin{table*}[!ht]
\caption{Few-shot classification accuracy ($\%$) on \textit{tiered}ImageNet. ({\textdagger}: only supervised setting is considered.)} 
\label{tab:classification_main_2}
    \small
    \centering
    \begin{small} 
    \begin{tabular}{l|cc|cc}
        \toprule
        & \multicolumn{2}{c|}{\textit{1-shot}} & \multicolumn{2}{c}{\textit{5-shot}}\\
        Methods & \textit{w/o OOD} & \textit{w/ OOD} & \textit{w/o OOD} & \textit{w/ OOD}\\
        \midrule
    
        Soft k-Means~\cite{ren2018meta}         &  27.53{\tiny $\pm$0.74} &	27.04{\tiny $\pm$0.76} &	44.63{\tiny $\pm$1.19} &	44.78{\tiny $\pm$1.05} \\
        Soft k-Means+Cluster~\cite{ren2018meta} &  30.48{\tiny $\pm$0.84} &	31.30{\tiny $\pm$0.86} &	46.93{\tiny $\pm$1.18} &	49.33{\tiny $\pm$1.17} \\
        Masked Soft k-Means~\cite{ren2018meta}  &  33.85{\tiny $\pm$0.84} &	32.99{\tiny $\pm$0.87} &	47.63{\tiny $\pm$1.12} &	47.35{\tiny $\pm$1.08} \\
        TPN-semi~\cite{liu2018learning} &  44.13{\tiny $\pm$1.04} &	31.83{\tiny $\pm$1.09} &	58.53{\tiny $\pm$1.57} &	56.92{\tiny $\pm$1.67} \\
        LST(\textit{small})~\cite{li2019learning}   & 42.86{\tiny $\pm$0.86}  & 42.33{\tiny $\pm$0.95}  &  59.55{\tiny $\pm$0.92}  & 58.82{\tiny $\pm$0.93}  \\
        LST(\textit{large})~\cite{li2019learning}   & 44.34{\tiny $\pm$0.97}  & 44.59{\tiny $\pm$0.99}  &  61.45{\tiny $\pm$0.90}  & 60.75{\tiny $\pm$0.93}  \\
        {MAML}{}\textsuperscript{\textdagger} \cite{finn2017model}     & 41.96{\tiny $\pm$0.84} &	41.96{\tiny $\pm$0.84} &	61.30{\tiny $\pm$0.85} &	61.30{\tiny $\pm$0.85} \\
        VAT \cite{miyato2018virtual}     &    41.52{\tiny $\pm$0.82} &	41.51{\tiny $\pm$0.79} &	59.98{\tiny $\pm$0.83} &	60.01{\tiny $\pm$0.87} \\
        PL \cite{lee2013pseudo}      &  41.22{\tiny $\pm$0.89} &	40.87{\tiny $\pm$0.83} &	61.70{\tiny $\pm$0.77}&	60.57{\tiny $\pm$0.87} \\ 
        \hline 
        \textsc{Gcmi} (ours)  &  45.49{\tiny $\pm$0.91} &	45.55{\tiny $\pm$0.90} &	63.67{\tiny $\pm$0.83} &	\textbf{62.59}{\tiny $\pm$0.85} \\
        \textsc{Flmi} (ours) &   \textbf{45.63}{\tiny $\pm$0.86} &	\textbf{46.19}{\tiny $\pm$0.94} &	\textbf{63.75}{\tiny $\pm$0.87} &	62.19{\tiny $\pm$0.91} \\
        \bottomrule
    \end{tabular}
    \end{small}
    \vspace{-1mm}
\end{table*}

\begin{table*}[!ht]
\caption{Few-shot classification accuracy ($\%$) on CIFAR-FS. ({\textdagger}: only supervised setting is considered.)} 
\label{tab:classification_main_3}
    \small
    \centering
    \begin{small} 
    \begin{tabular}{l|cc|cc}
        \toprule
        & \multicolumn{2}{c|}{\textit{1-shot}} & \multicolumn{2}{c}{\textit{5-shot}}\\
        Methods & \textit{w/o OOD} & \textit{w/ OOD} & \textit{w/o OOD} & \textit{w/ OOD}\\
        \midrule
        LST(\textit{small})~\cite{li2019learning}   & 38.60{\tiny $\pm$0.94}  & 38.61{\tiny $\pm$0.89}  &  53.43{\tiny $\pm$0.94}  & 51.83{\tiny $\pm$0.98}  \\
        LST(\textit{large})~\cite{li2019learning}   & 37.65{\tiny $\pm$0.91}  & 37.74{\tiny $\pm$0.97}  &  \textbf{55.77}{\tiny $\pm$0.93}  & 52.23{\tiny $\pm$0.95}  \\

        {MAML}{}\textsuperscript{\textdagger} \cite{finn2017model}    & 37.90{\tiny $\pm$0.91} &	37.90{\tiny $\pm$0.91} &	52.60{\tiny $\pm$0.89} &	52.60{\tiny $\pm$0.89} \\
        VAT \cite{miyato2018virtual}    &  39.48{\tiny $\pm$0.83} &	38.91{\tiny $\pm$0.88} &	53.20{\tiny $\pm$0.80} &	52.44{\tiny $\pm$0.83} \\
        PL \cite{lee2013pseudo}         &  38.11{\tiny $\pm$0.87} &	37.29{\tiny $\pm$0.92} &	52.83{\tiny $\pm$0.82} &	52.42{\tiny $\pm$0.91} \\
        \hline
        \textsc{Gcmi} (ours)  &   40.47{\tiny $\pm$0.88} &	40.10{\tiny $\pm$0.89} &	{55.01}{\tiny $\pm$0.84} &	\textbf{54.42}{\tiny $\pm$0.92} \\
        \textsc{Flmi} (ours)  &   \textbf{40.96}{\tiny $\pm$0.86} &	\textbf{40.48}{\tiny $\pm$0.87} &	54.94{\tiny $\pm$0.80} &	54.16{\tiny $\pm$0.91}  \\
        \bottomrule
    \end{tabular}
    \end{small}
    \vspace{-3mm}
\end{table*}

\subsection{Scalability of SMI Optimization} \label{sec:smi_scalability}
We chose to embed semi-supervision using \textsc{Flmi} and \textsc{Gcmi} in our framework due to their scalability benefits \cite{kothawade2021similar, kothawade2021talisman}. Asymptotically, the time and space complexity of computing a similarity matrix $\Xcal$ for \textsc{Flmi} and \textsc{Gcmi} is only $|\Rcal| \times |\Ucal|$. Since in the few-shot learning setting, we set $\Rcal \leftarrow {\Scal \cup \Qcal}$ which is comparatively much smaller than $\Ucal$, the complexity of these SMI functions is only $|\Ucal|$. Moreover, the SMI functions that we use are monotone and submodular which allows a $1 - \frac{1}{e}$ constant factor approximation \cite{nemhauser1978analysis}. Hence, for optimizing the SMI functions, we use a greedy algorithm \cite{nemhauser1978analysis} using memoization \cite{iyer2019memoization}. This leads to an amortized cost of $|\Ucal|\log|\Ucal|$ which can be further reduced to $|\Ucal|$ using the lazier than lazy greedy algorithm \cite{mirzasoleiman2015lazier}. Hence, \textsc{Flmi} and \textsc{Gcmi} can be optimized in linear time, making it applicable to few-shot learning datasets with a large number of tasks and large unlabeled sets.

\textbf{Time complexity of \textsc{Platinum}.} Since \textsc{Flmi} and \textsc{Gcmi} can be optimized in linear time, asymptotically it does not change in terms of the worst case in MAML. Therefore, the iteration complexity is still $\mathcal{O}(1/\epsilon^2)$~\cite{fallah2020convergence}. \tabref{tab: running_time} shows the empirical running time per epoch (100 iterations, one task per iteration) for MAML and our proposed \textsc{Gcmi} and \textsc{Flmi}. 
Therefore, it is safe to say that our proposed framework \model{} would not slow down the original meta-learning framework (such as MAML).     

\section{Experiments} \label{sec:experiments}

In this section, we evaluate \model\ for semi-supervised few-shot image classification by comparing the accuracy attained at meta-testing. In \secref{sec:main_results}, we compare our method with the state-of-the-art techniques on a diverse set of datasets and settings. In \secref{sec:ablation_results}, we discuss multiple ablation studies by varying the number OOD classes in the unlabeled set and studying the effect of the proposed semi-supervision in the inner \emph{and} outer loop.   

In order to demonstrate the effectiveness of \model{}, we aim to study two questions:

\textbf{Q1}: Can \model{} be successfully applied to semi-supervised few shot classification scenario with very few labeled examples on the top of MAML and boost the performance of MAML with the additional unlabeled set?\\
\textbf{Q2}: In realistic scenarios, the unlabeled set is bound to have OOD data. Can \model{} provide robust semi-supervision by ignoring the OOD data in the unlabeled set?


\subsection{Datasets and Implementation details} \label{sec:implementation_details}

\textbf{Datasets.} We conduct experiments on three datasets: \textit{mini}ImageNet \cite{vinyals2016matching}, \textit{tiered}ImageNet \cite{ren2018meta}, and CIFAR-FS \cite{bertinetto2018meta}. 
Both \textit{mini}ImageNet and \textit{tiered}ImageNet are modified subsets of the ILSVRC-12 dataset~\cite{russakovsky2015imagenet}. \textbf{\textit{mini}ImageNet} consists of 100 classes and each class has 600 images. Following the disjoint class split from ~\cite{ravi2016optimization}, we split it into 64 classes for training, 16 for validation, and 20 for test. Similarly, \textbf{\textit{tiered}ImageNet} is a larger dataset, consisting of 608 classes and each class has 768$\sim$1300 images. Classes are split into 351 for training, 97 for validation, and 160 for test~\cite{ren2018meta}. All images in these two datasets are of resolution 84$\times$84$\times$3.
\textbf{CIFAR-FS} contains 60,000 images of size 32$\times$32$\times$3 from 100 classes. We use the same class split as \textit{mini}ImageNet.

\textbf{Implementation details.} We follow the ``$K$-shot, $M$-way'' episode training setting in \cite{finn2017model} and~\cite{ren2018meta} to do semi-supervised few-shot classification experiments to evaluate \model{}. In all our experiments, we use first-order approximation of the MAML algorithm \cite{finn2017model}.  
We implement image classification experiments in 5-way, 1-shot (5-shot) settings. Concretely, at first, all examples of each class will be randomly divided into labeled portion (where $\Scal$ and $\Qcal$ are sampled from) and unlabeled portion (where $\Ucal$ is sampled from) based on a predefined labeled ratio $\rho$, where $\rho$ is the ratio of the number of data points in the labeled portion to the total number of data points in the current class. Then, we sample each task to contain 1 (5) data points in the support set $\Scal$, and 15 (15) data points in the query set $\Qcal$ per class. For the unlabeled set $\Ucal$, we sample 50 (50) data points for each class. To select a subset for semi-supervision using SMI functions, we use a budget $B_{in}=$ 25 (25) for the inner loop, and a budget $B_{out}=$ 50 (50) for the outer loop.  Note that we perform a per-class selection to assign pseduo-labels using the SMI functions, which leads to a budget of 5 and 10 data points for the inner and outer loop respectively. For our experiments in \tabref{tab:classification_main_1}, \tabref{tab:classification_main_2} and \tabref{tab:classification_main_3}, we use a labeled set ratio $\rho=0.01$. However, we also compare with a number of other $\rho$ values (see \tabref{tab:classification_main_ratio} and \tabref{tab:add_orig_40}). For our experiments with OOD classes in the unlabeled set (\tabref{tab:classification_main_2}), we use 5 distractor classes with 50 data points for each class. To make a fair comparison, we apply the same 4-layer CONV backbone architecture given in~\citep{vinyals2016matching, finn2017model} for our model and all baselines. We provide detailed hyperparameters for our experiments in Appendix \ref{app:exp_details}. We use an NVIDIA RTX A6000 GPU for our experiments. The PyTorch implementation is available at \url{https://github.com/Hugo101/PLATINUM}.

\textbf{Baselines.} We consider meta-learning based semi-supervised few-shot classification approaches as baselines, including the extended prototypical network~\cite{ren2018meta} (including Soft k-Means, Soft k-Means+Cluster, Masked Soft k-Means), TPN-semi~\cite{liu2018learning}, and reimplemented LST~\cite{li2019learning} based on the 4-layer CONV (\textit{small}/\textit{large}:small/large inner loop steps, {\em i.e.} 5/10 (meta-train/test) for \textit{small}, 20/100 for \textit{large}.). We also compare with MAML which serves as the supervised classification baseline without the additional unlabeled set. In addition, we compare with two well known approaches from the semi-supervised learning literature and implement them in the inner and outer loop on the top of MAML. The first one is Pseudo-labeling (PL)~\cite{lee2013pseudo} and the second one is a consistency regularization method, VAT~\cite{miyato2018virtual}.

\begin{table*}[!htbp]
\caption{Few-shot classification accuracy (\%) of different labeled ratios ($\rho=$1\%, 5\%, 10\%, 20\%, 30\%) on \textit{mini}ImageNet. (Due to the space limitation, we show results with 95$\%$ confidence interval in Appendix~\ref{app:exp_details}. {\textdagger}: only supervised setting is considered.)} 
\label{tab:classification_main_ratio}
    \small
    \vspace{-2mm}
    \centering
    \begin{small} 
    \begin{tabular}{l|ccccc|cccc}
        \toprule
        & \multicolumn{5}{c}{\textit{1-shot}} & \multicolumn{4}{c}{\textit{5-shot}} \\
        Methods & 1\% & 5\% & 10\% & 20\% & 30\% & 1\% & 10\% & 20\% & 30\% \\
        \midrule
        
        Soft k-Means~\cite{ren2018meta}         &  24.61 &	38.45 &	40.65 &	42.55 &	44.09 &	38.20 &	56.27 &	60.13 &	62.47 \\
        Soft k-Means+Cluster~\cite{ren2018meta} &  15.76 &	38.34 &	41.15 &	45.17 &	47.05 &	33.65 &	56.87 &	60.33 &	62.43 \\
        Masked Soft k-Means~\cite{ren2018meta}  &  25.48 &	39.03 &	42.91 &	45.31 &	47.17 &	39.33 &	57.20 &	62.50 &	63.00 \\
        TPN-semi~\cite{liu2018learning}         &  40.25 &	42.40 &	45.78 &	48.02 &	47.52 &	46.27 &	60.55 &	62.43 &	63.10 \\
        MAML{}\textsuperscript{\textdagger}~\cite{finn2017model}               &  35.26 &	42.51 &	44.29 &	45.10 &	45.26 &	60.22 &	61.06 &	63.18 &	65.60 \\
        PL~\cite{lee2013pseudo}                 &  37.71 &	44.04 &	46.58 &	45.13 &	44.37 &	60.64 &	61.17 &	63.06 &	65.14 \\
        \hline
        \textsc{Gcmi} (ours)                    & 41.94 &	44.98 &	46.85 &	47.72 &	48.93 &	63.62 &	\textbf{62.72} &	64.78 &	65.96 \\
        \textsc{Flmi} (ours)                    & \textbf{42.27} &	\textbf{45.01} &	\textbf{47.84} &	\textbf{47.82} &	\textbf{48.98} &	\textbf{63.80} &	62.60 &	\textbf{65.16} &	\textbf{66.10} \\
        \bottomrule
    \end{tabular}
    \end{small}
\end{table*}

\begin{table*}[!ht]
\caption{\small Few-shot classification accuracy (\%) with $\rho=$40\% labeled data per class for \textit{mini}ImageNet.}
\label{tab:add_orig_40}
    \small 
    \centering
    \begin{small}
    \begin{tabular}{l | c c |c c }
        \toprule
         & \multicolumn{2}{c|}{\textit{1-shot}} & \multicolumn{2}{c}{\textit{5-shot}}\\
        Methods  & \textit{w/o OOD} & \textit{w/ OOD}  & \textit{w/o OOD} & \textit{w/ OOD}  \\
        \midrule
        Soft k-Means~\cite{ren2018meta}          & 50.09{\tiny $\pm$0.45}  & 48.70{\tiny $\pm$0.32}  & 64.59{\tiny $\pm$0.28}  & 63.55{\tiny $\pm$0.28}  \\
        Soft k-Means Cluster~\cite{ren2018meta}  & 49.03{\tiny $\pm$0.24}  & 48.86{\tiny $\pm$0.32}  & 63.08{\tiny $\pm$0.18}  & 61.27{\tiny $\pm$0.24}  \\
        Masked Soft k-Means~\cite{ren2018meta}   & 50.41{\tiny $\pm$0.31}  & 49.04{\tiny $\pm$0.31}  & 64.39{\tiny $\pm$0.24}  & 62.96{\tiny $\pm$0.14}  \\
        TPN-semi~\cite{liu2018learning}              & \textbf{52.78}{\tiny $\pm$0.27}  & \textbf{50.43}{\tiny $\pm$0.84}  & 66.42{\tiny $\pm$0.21}  & 64.95{\tiny $\pm$0.73}  \\
        \hline 
        \textsc{Gcmi} (\textit{large}, ours)  & \textbf{51.35}{\tiny $\pm$0.93} & \textbf{50.85}{\tiny $\pm$0.89} & \textbf{66.65}{\tiny $\pm$0.75} & \textbf{66.66}{\tiny $\pm$0.74}  \\
        \textsc{Flmi} (\textit{large}, ours)  & 51.06{\tiny $\pm$0.96} & 49.83{\tiny $\pm$0.91}       & \textbf{67.34}{\tiny $\pm$0.72} & \textbf{66.20}{\tiny $\pm$0.73} \\
        \bottomrule 
    \end{tabular}
\vspace{-3mm}
\end{small}

\end{table*}
\subsection{Results} \label{sec:main_results}
In this section, we present extensive experiments that compare the performance of \model\ with other methods. We provide the results for 1-shot (5-shot), 5-way experiments for \textit{mini}ImageNet in \tabref{tab:classification_main_1}, \textit{tiered}ImageNet in \tabref{tab:classification_main_2}, and CIFAR-FS in \tabref{tab:classification_main_3}. On all datasets, we conduct experiments with (\textit{w/}) and without (\textit{w/o}) OOD classes in the unlabeled set. Since these experiments use $\rho=0.01$, we conduct experiments for $\rho=$0.1, 0.2 and 0.3 on \textit{mini}ImageNet, and present the results in \tabref{tab:classification_main_ratio}. \tabref{tab:add_orig_40} shows the results of $\rho=$0.4 on \textit{mini}ImageNet which is the default setting in recent papers.

\textbf{Analysis across multiple datasets.} We observe that \textsc{Flmi} outperforms other methods for the 1-shot setting on all datasets. The performance of \textsc{Flmi} is slightly better than \textsc{Gcmi} due to the additional diversity that the \textsc{Flmi} function models (see \tabref{tab:submod_inst}). When compared to other methods, the SMI functions (\textsc{Flmi} and \textsc{Gcmi}) improve the accuracy by $\approx 2 - 4\%$ over existing methods. Interestingly, in \tabref{tab:classification_main_3}, we observe that \textsc{Gcmi} outperforms \textsc{Flmi} and other baselines in the presence of OOD classes in the unlabeled set. This is expected since \textsc{Gcmi} \emph{only} models query-relevance \cite{kothawade2021prism} as opposed to \textsc{Flmi} which also models diversity.

\textbf{Varying the labeled set ratio $\rho$. } In \tabref{tab:classification_main_ratio}, we analyze different values of $\rho$ for 1-shot and 5-shot on the \textit{mini}ImageNet dataset. We observe that the gain using \model\ is higher when the number of labeled data points is lower than the number of unlabeled data points, \ie, $\rho$ is small. This further reinforces the need for a framework like \model\ which performs well in the low labeled data regime. It should be noted that, in our experiments, we use a much lower labeled ratio $\rho=$0.01, since that is a more realistic scenario. This could explain why the results of many baselines look low. We also conduct an experiment with the exact setup as~\cite{ren2018meta} where they assume labels for 40\% of examples per class for \textit{mini}ImageNet. We present the results in \tabref{tab:add_orig_40} and observe that our results still outperform~\citet{ren2018meta} and~\citet{liu2018learning}'s for 5-shot case at least. As mentioned by~\citet{ren2018meta}, their method \textit{overfits} when the labeled data is as low as $\rho=$0.1 (\textit{mini}ImageNet), whereas our method works still well in the low data regime as well.


\begin{figure}[ht]
\vskip 0.2in
\begin{center}
\centerline{\includegraphics[width=0.9\columnwidth]{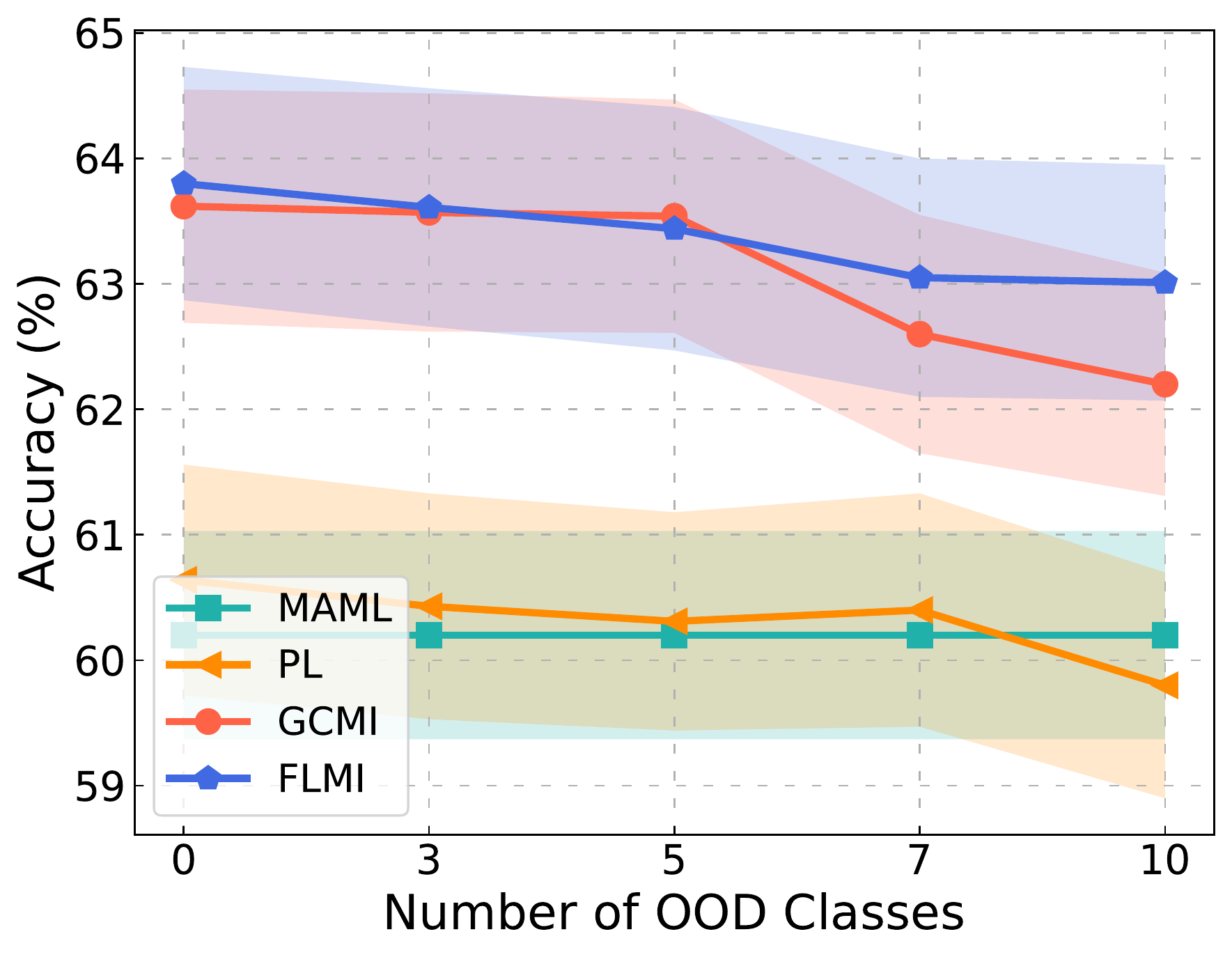}}
\caption{Comparison under different number of OOD classes in the Unlabeled Set for 5-shot case on \textit{mini}ImageNet. TPN-semi is much worse than MAML by 20$\%$, so we do not put it in this figure. (1-shot case is shown in Appendix~\ref{app:exp_details}.)}
\label{fig:num_OOD_classes_5shot}
\end{center}
\vskip -0.2in
\end{figure}

\begin{figure}[ht]
\vskip 0.2in
\begin{center}
\centerline{\includegraphics[width=\columnwidth]{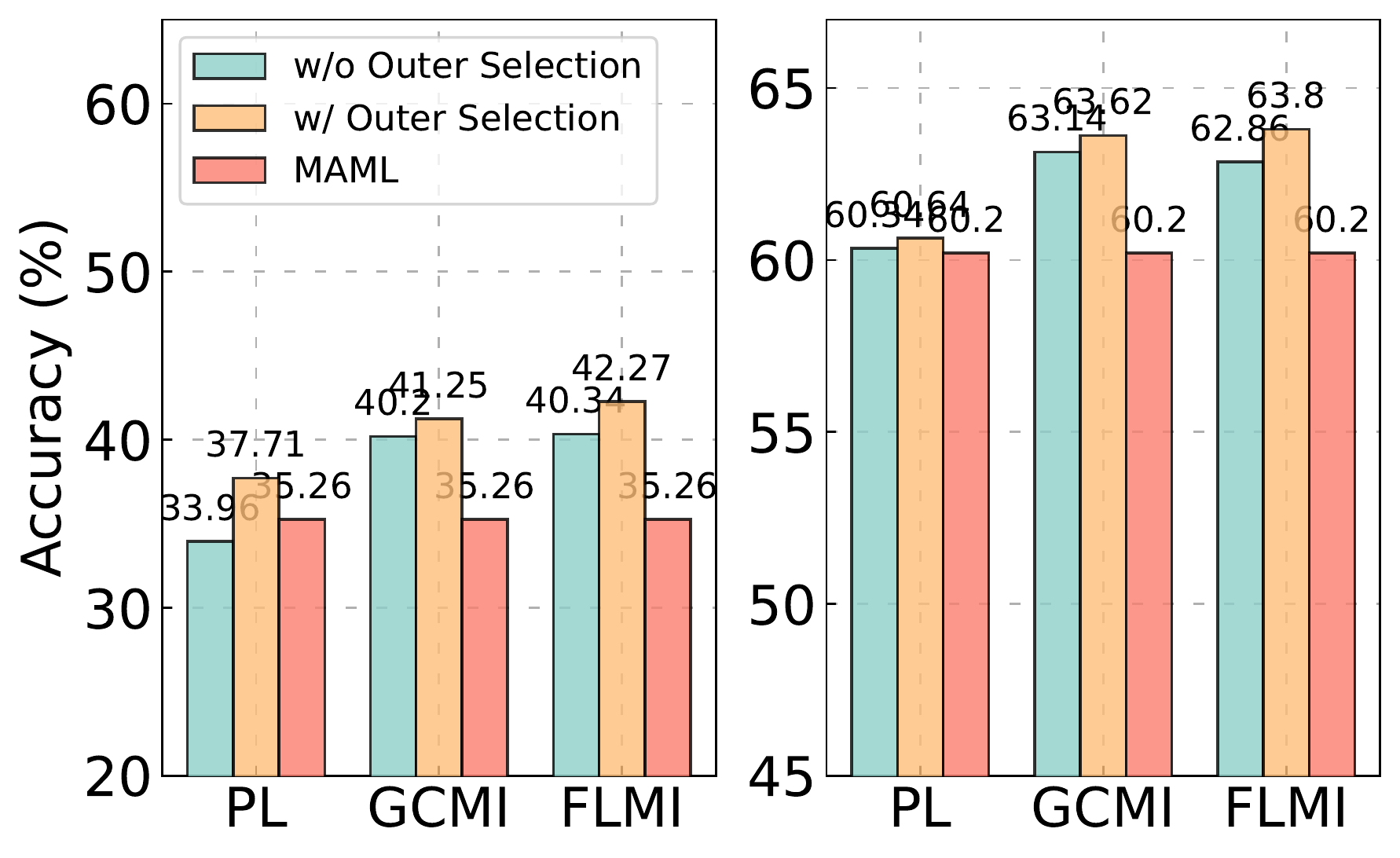}}
\caption{\textit{w/} outer selection vs. \textit{w/o} outer selection. \textbf{Left}: 1-shot, \textbf{Right}: 5-shot. Both of them are on \textit{mini}ImageNet.}
\label{fig:inner_outer_select}
\end{center}
\vskip -0.2in
\end{figure}


\subsection{Ablation Study} \label{sec:ablation_results}

\textbf{Varying the number of distractor classes.} To explore the effect of the number of OOD classes in the unlabeled set, we increased the number of OOD classes, while keeping the number of in-distribution classes to be 5. We keep using the same number of unlabeled images per class as previous experiment. In \figref{fig:num_OOD_classes_5shot}, we present the result for this ablation study on the 5-way 5-shot setting for the \textit{mini}ImageNet dataset. As expected, we observe that the accuracy during meta-testing decreases as the complexity of in-distribution subset selection increases due to larger number of OOD classes. We observe that the semi-supervision provided by the SMI based methods (\textsc{Flmi} and \textsc{Gcmi}) consistently aids MAML and outperform other methods as the number of OOD classes increase, while PL suffers and eventually performs slightly worse than MAML.

\textbf{Inner and outer loop selection.}
One of the key components of \model\ is embedding semi-supervision in the outer-loop. We conduct an ablation study using the 5-way 1-shot and 5-shot setting on the \textit{mini}ImageNet dataset to analyze the effect of outer-loop semi-supervision and present the results in \figref{fig:inner_outer_select}. Particularly, we evaluate the meta-test accuracy of few-shot classification with (\textit{w/}) and without (\textit{w/o}) the outer-loop selection for three methods: PL, \textsc{Gcmi} and \textsc{Flmi}. MAML is also included for comparison. We observe that providing semi-supervision in the outer loop consistently improves the performance across all experiments. Especially for 1-shot of PL, we observe an improvement of $\approx4\%$. Interestingly, PL performs worse than MAML without the outer-loop semi-supervision, and outperforms MAML with it.

\begin{table}[t!]
\caption{\small The accuracy (\%) of 5-way 5-shot experiment for $\rho=$0.4 on \textit{mini}ImageNet based on a pretrained ResNet-12 (\textit{w/o OOD}).}
\label{tab:add_resnet}
    \small 
    \vspace{-1mm}
    \centering
    \begin{center}
    \begin{tabular}{c c c }
        \toprule
        MAML & LST~\cite{li2019learning} & \textsc{Gcmi} (\textit{large}, ours)\\
        \midrule
        75.21{\tiny $\pm$0.65}  & 78.70{\tiny $\pm$0.80}  & 79.44{\tiny $\pm$0.76}  \\
    
    \bottomrule 
    \end{tabular}
\vspace{-3mm}
\end{center}
\vspace{-2mm}
\end{table}

\textbf{Other backbones.} We primarily use the 4-layer CONV architecture so that we could do a fair comparison with existing methods~\cite{ren2018meta,liu2018learning} which have a similar setting as ours. The more recent semi-supervised few-shot learning and meta-learning methods~\cite{li2019learning, tian2020rethinking} use ResNet-12 as backbone to boost the performance. Note that they usually require a feature pretraining procedure, which may not be available in various real-world scenarios, especially in the case of low labeled ratio in our paper. To compare, we conduct experiment based on a pretrained ResNet-12 from~\citet{tian2020rethinking} to replace 4-layer CONV, and only update the final classifier. We report results in \tabref{tab:add_resnet} and observe that SMI could improve further. It is reasonable that using a more powerful feature extractor model would still result in gains over current methods.



\section{Conclusion}
In this paper, we propose a novel semi-supervised model-agnostic meta-learning framework \model{}. It leverages submodular mutual information functions as per-class acquisition functions to select more balanced and diverse data from unlabeled data in the inner and outer loop of meta-learning. Meta-learning based semi-supervised few-shot learning experiments validates the effectiveness of embedding semi-supervision in the MAML by \model{}, especially for small ratio of labeled to unlabeled samples. We also notice that it might be useful to involve some diversity measurements for the selected subset to do quantitative analysis, we leave this as future work.

\section*{Acknowledgements}
We gratefully thank anonymous reviewers for their valuable comments. This work is supported by the National Science Foundation under Grant Numbers IIS-2106937, IIS-1954376, IIS-1815696, a gift from Google Research, and the Adobe Data Science Research award. Any opinions, findings, and conclusions or recommendations expressed in this material are those of the authors and do not necessarily reflect the views of the National Science Foundation, Google or Adobe.



\bibliographystyle{icml2022}
\bibliography{main}

\newpage
\appendix
\onecolumn

\section{Notation}
\label{app:notations}
For clear interpretation, we list the notations used in this paper and their corresponding explanation, as shown in Table~\ref{tab:notation}.

{\renewcommand{\arraystretch}{1.2} 
\begin{table}[ht!]
\caption{Important Notations and Descriptions} 
\label{tab:notation}
    \centering
\vskip 0.15in    
    \begin{center}
    \begin{small} 
    \begin{tabular}{ll}
        \toprule
        \bfseries{Notation} & \bfseries{Description}  \\
        \midrule
        $p(\mathcal{T})$ & Probability distribution of meta-training tasks  \\
        $N$ & The number of meta-training tasks\\
        $M$-way, $K$-shot & The number of classes in one task $M$, and each class contains $K$ examples in the support set \\
        $\mathcal{T}_i$ & The $i$-th meta-training task \\
        $\{\mathcal{S}_i$, $\mathcal{Q}_i$, $\mathcal{U}_i$\} & Support set, query set, and unlabeled set of meta-training task $\mathcal{T}_i$ \\
        \{$\mathcal{S}^{new}$, $\mathcal{Q}^{new}$, $\mathcal{U}^{new}$\} & Support set, query set, and unlabeled set for task $\mathcal{T}'_i$ in meta-test \\
        $\mathcal{A}^{s}_{i}$ & Selected unlabeled examples from Unlabeled set for task $\mathcal{T}_i$ in the inner loop in meta-training \\
        $\mathcal{A}^{q}_{i}$ & Selected unlabeled examples from Unlabeled set for task $\mathcal{T}_i$ in the outer loop in meta-training \\
        $\theta$  & Initial parameters of base learner \\
        $\phi_i$  & Task-specific parameters for task $\mathcal{T}_i$ \\
        ${L}(\phi; \mathcal{D})$ & Loss function on dataset $\mathcal{D}$ characterized by model parameter $\phi$ \\
        $L_{l}$, $L_{u}$ & Cross entropy loss on labeled data ( or hypothesized labeled data) \\
        $\text{Alg}(\theta; \mathcal{D})$ & One or multiple steps of gradient descent initialized at $\theta$ on dataset $\mathcal{D}$\\
        $\alpha, \beta$ & Learning rate in the inner loop and outer loop  \\
        $\tau_{\text{in}}$, $\tau_{\text{out}}$ & Temperature annealing coefficient in the inner (or outer) loop \\
        
        $B_{\text{in}}$, $B_{\text{out}}$ & Budget in the inner (or outer) loop selection among all classes in the task\\
        $T_{\text{in}}$, $T_{\text{out}}$  & Total number of steps in the inner loop; The number of epochs in the outer loop; \\
        $T_{\text{warm}}$ & Warm start epoch in the outer loop \\
        $f$ & A submodular function\\
        $S_{ij}$ & similarity between sample $i$ and $j$ \\
        $I_f$ & A submodular mutual information function instantiated using a submodular function $f$\\
        $\Xcal$ & Pairwise similarity matrix used to instantiate a submodular function $f$\\
        
        \bottomrule
    
    \end{tabular}
    \end{small}
\end{center}

\end{table}
}

\section{Details of Inner and Outer SMI Subset Selection} \label{app:inner_outer}
\subsection{Inner loop}

Task specific model parameters for task $\mathcal{T}_i$: 
\begin{equation}
    \begin{aligned}
\phi_{i}=& \operatorname{argmin}_{\theta} L_{l}\left(\theta ; \mathcal{S}_{i}\right)+\tau_{\text{in}} L_{u}\left(\theta ; \mathcal{A}^{s}_{i}\right) \\
=&\theta-\alpha \nabla_{\theta} L_{l}\left(\theta ; \mathcal{S}_{i}\right)-\alpha \tau_{\text{in}} \nabla_{\theta} L_{u}\left(\theta ; \mathcal{A}^{s}_{i}\right)  \text{(one step gradient update example)}
\end{aligned}
\end{equation}

In which, $\alpha$ is learning rate. $\tau$ is the coefficient from the pseudo labeling approach. Since it is an increasing temperature variable, let $\tau^{(t)}$ denote the $\tau$ in step $t$.

Since there are several gradient update steps in inner loop. 

\begin{equation}
    \phi_i^{(t+1)} = \phi_i^{(t)} - \nabla  L\left(\theta ; \mathcal{S}_{i}\cup\mathcal{A}^{s}_{i}\right)
\end{equation}

Let $\phi_{i}^{(t)}$ denote the model parameters for $t$-th step for task $\mathcal{T}_i$.
\begin{itemize}
    \item initialization: $\phi_{i}^{(0)} = \theta$, $\mathcal{A}^s_i=\emptyset$  
    \item inner step 1: 
    $$
\phi_i^{(1)}=\theta-\alpha \nabla_{\theta} L_{l}(\theta ; \mathcal{S}_i)-\alpha \tau_{\text{in}}^{(1)} \nabla_{\theta} L_{u}\left(\theta ; \mathcal{A}^{s}_{i}\right)
$$

Select subset for this step:
$\mathcal{A}^{s}_{i1} \leftarrow \operatorname{argmax}_{\mathcal{A}_{i1}^s \subseteq \mathcal{U}_i,|\mathcal{A}^s_{i1}| \leq B_{\text{in}}} I_{f}(\mathcal{A}^s_{i1}; \mathcal{S}_i \cup \mathcal{Q}_i)$

Set of selected examples: $\mathcal{A}^s_i$ = $\mathcal{A}^s_i \cup \mathcal{A}^s_{i1}$

In this step, the CNN model used to calculate the class probabilities for SMI is parameterized by {\color{blue}$\theta$}.

    \item inner step 2: 
     $$
\phi_i^{(2)}=\theta-\alpha \nabla_{\phi^{(1)}} L_{l}(\phi^{(1)} ; \mathcal{S}_i)-\alpha \tau_{\text{in}}^{(2)} \nabla_{\phi^{(1)}} L_{u}\left(\phi^{(1)} ; \mathcal{A}^{s}_{i}\right)
$$

Select subset for this step:
$\mathcal{A}^{s}_{i2} \leftarrow \operatorname{argmax}_{\mathcal{A}_{i2}^s \subseteq \mathcal{U}_i,|\mathcal{A}^s_{i2}| \leq B_{\text{in}}} I_{f}(\mathcal{A}^s_{i2}; \mathcal{S}_i \cup \mathcal{Q}_i)$

Set of selected examples: $\mathcal{A}^s_i$ = $\mathcal{A}^s_i \cup \mathcal{A}^s_{i2}$

In this step, the CNN model used to calculate the class probabilities for SMI is parameterized by {\color{blue}$\phi^{(1)}$}.

    \item inner step 3: 
     $$
\phi_i^{(3)}=\theta-\alpha \nabla_{\phi^{(2)}} L_{l}(\phi^{(2)} ; \mathcal{S}_i)-\alpha \tau_{\text{in}}^{(3)} \nabla_{\phi^{(2)}} L_{u}\left(\phi^{(2)}; \mathcal{A}^{s}_{i}\right)
$$

Select subset for this step:
$\mathcal{A}^{s}_{i3} \leftarrow \operatorname{argmax}_{\mathcal{A}_{i3}^s \subseteq \mathcal{U}_i,|\mathcal{A}^s_{i3}| \leq B_{\text{in}}} I_{f}(\mathcal{A}^s_{i3}; \mathcal{S}_i \cup \mathcal{Q}_i)$

Set of selected examples: $\mathcal{A}^s_i$ = $\mathcal{A}^s_i \cup \mathcal{A}^s_{i3}$

In this step,the CNN model used to calculate the class probabilities for SMI is parameterized by {\color{blue}$\phi^{(2)}$}.

    \item continue repeat until the end of inner loop: step  $T_{\text{in}}-1$.
    
    Model parameters for task $T_i$ udpate process:
    
    $$\phi^{(0)} (:=\theta) \rightarrow \phi^{(1)}\rightarrow \phi^{(2)}\rightarrow \phi^{(3)} ...\rightarrow \phi^{(T_{\text{in}}-1)} $$
    
\end{itemize}

\subsection{Outer Loop}
Meta-parameter update according to:
\begin{equation}
    \theta = \argmin_{\theta} \mathcal{J}(\theta)
\end{equation}

The final loss function is:

\begin{equation}
\begin{aligned}
\mathcal{J}(\theta)&=\frac{1}{M}\sum_{i=1}^{M} L_{l}\left(\phi_{i} ; \mathcal{Q}_{i}\right)+ \tau_{\text{out}} L_{u}\left(\phi_{i} ; \mathcal{A}^{q}_{i}\right) \\
&=\frac{1}{M}\sum_{i=1}^{M} L_{l}\left(\operatorname{argmin}_{\theta} L_{l}\left(\theta ; \mathcal{S}_{i}\right)+\tau_{\text{in}} L_{u}\left(\theta ; \mathcal{A}^{s}_{i}\right) ; \mathcal{Q}_{i}\right)+ \tau_{\text{out}} L_{u}\left(\operatorname{argmin}_{\theta} L_{l}\left(\theta ; \mathcal{S}_{i}\right)+\tau_{\text{in}} L_{u}\left(\theta ; \mathcal{A}^{s}_{i}\right) ; \mathcal{A}^{q}_{i}\right)
\end{aligned}
\end{equation}

in which, $T$ is the set of meta-training tasks. $\tau_{\text{out}}$ is still a coefficient borrowed from the pseudo label formulation.

The second equal in the above equation is according to the inner loop update:
$$
\phi_i = \operatorname{argmin}_{\theta} L_{l}\left(\theta ; \mathcal{S}_{i}\right)+\tau L_{u}\left(\theta ; \mathcal{A}^{s}_{i}\right)
$$

Since there is only one step in the outer loop, subset selection only happens one time.

\begin{equation}
     \mathcal{A}^{q}_{i} \leftarrow \argmax_{\mathcal{A}\subseteq \mathcal{U}_i\backslash\mathcal{A}^s_i, |\mathcal{A}|\leq B_{\text{out}}}I_{f}(\mathcal{A}; \mathcal{S}_i\cup \mathcal{Q}_i)
\end{equation}
The CNN model used here should be the final model parameter after the inner loop: {\color{blue}$\phi_i^{(T_{\text{in}}-1)}$}.

The motivation here to use $\phi_i^{(T_{\text{in}}-1)}$ instead of meta-parameter $\theta$ is that SMI needs a CNN model which has powerful representation. For task $\mathcal{T}_i$, $\phi_i^{(T_{\text{in}}-1)}$ is more powerful than $\theta$.







\section{Additional Experimental Detail}
\label{app:exp_details}


As aforementioned, our backbone follows the same architecture as the embedding function used by \citep{finn2017model}. Concretely, the backbone structure is made of 4 modules, each of which contains a 3$\times$3 convolutions and 64 filters, followed by batch normalization, a ReLU, and a 2$\times$2 max-pooling with stride 2. To reduce overfitting, 32 filters per layer are considered. Cross entropy loss function is used for all experiment including the loss of selected unlabeled set between the predicted labels and the hypothesized labels.

\subsection{Hyparameters tuning} 
All baseline approaches including three extended prototypical networks~\cite{ren2018meta} and TPN-semi~\cite{liu2018learning} are reimplemented via their official code following the original implementation including hyper-parameters. For our \model{} algorithm, all step sizes ($\alpha, \beta$) are chosen from $\{$0.0001, 0.001, 0.01, 0.1$\}$. The Batch size (number of tasks per iteration) is chosen from $\{$1, 2, 4$\}$. The number of iterations are chosen from $\{$10,000,\enspace 20,000,\enspace 30,000,\enspace 40,000,\enspace 60,000$\}$. The selected best ones are: learning rate in the inner loop $\alpha=0.01$, meta parameters step size (outer learning rate) $\beta=0.0001$; the number of iterations for all experiments is set to be 60,000 (600 epochs, each epoch has 100 iterations). We monitor the accuracy and loss from meta-validation stage and save the model which has the best validation accuracy and use that to evaluate the performance on unseen novel tasks in meta-test stage.

\subsection{Additional Results}

\begin{table*}[!htbp]
\caption{1-shot classification accuracies ($\%$) of different labeled ratios on \textit{mini}ImageNet.} 
\label{app:tab:classification_main_ratio_1_shot}
    \small
    \vspace{-3mm}
    \centering
    \begin{small} 
    \begin{tabular}{l|ccccc}
        \toprule
        & \multicolumn{5}{c}{\textit{1-shot}}  \\
        Methods & 1\% & 5\% & 10\% & 20\% & 30\%  \\
        \midrule
       
        Soft k-Means~\cite{ren2018meta}         & 24.61{\tiny $\pm$0.64} &	38.45{\tiny $\pm$0.81} &	40.65{\tiny $\pm$0.92} &	42.55{\tiny $\pm$0.99} &	44.09{\tiny $\pm$1.08} \\
        Soft k-Means+Cluster~\cite{ren2018meta} & 15.76{\tiny $\pm$0.59} &	38.34{\tiny $\pm$0.82} &	41.15{\tiny $\pm$0.99} &	45.17{\tiny $\pm$0.95} &	47.05{\tiny $\pm$1.08} \\
        Masked Soft k-Means~\cite{ren2018meta}  & 25.48{\tiny $\pm$0.67} &	39.03{\tiny $\pm$0.89} &	42.91{\tiny $\pm$0.93} &	45.31{\tiny $\pm$1.01} &	47.17{\tiny $\pm$1.07}  \\
        TPN-semi~\cite{liu2018learning}             & 40.25{\tiny $\pm$0.92} &	42.40{\tiny $\pm$0.77} &	45.78{\tiny $\pm$0.80} &	48.02{\tiny $\pm$0.82} &	47.52{\tiny $\pm$0.83} \\
         MAML~\cite{finn2017model}                & 35.26{\tiny $\pm$0.85} &	42.51{\tiny $\pm$0.78} &	44.29{\tiny $\pm$0.78} &	45.10{\tiny $\pm$0.75} &	45.26{\tiny $\pm$0.78} \\
        PL                   & 37.71{\tiny $\pm$0.94} &	44.04{\tiny $\pm$0.82} &	46.58{\tiny $\pm$0.72} &	45.13{\tiny $\pm$0.78} &	44.37{\tiny $\pm$0.81} \\
        \hline 
        \textsc{Gcmi} (ours)       & 41.94{\tiny $\pm$0.96} &	44.98{\tiny $\pm$0.80} &	46.85{\tiny $\pm$0.74} &	47.72{\tiny $\pm$0.76} &	48.93{\tiny $\pm$0.70}  \\
        \textsc{Flmi} (ours)       & \textbf{42.27}{\tiny $\pm$0.95} &	\textbf{45.01}{\tiny $\pm$0.83} &	\textbf{47.84}{\tiny $\pm$0.86} &	\textbf{47.82}{\tiny $\pm$0.78} &	\textbf{48.98}{\tiny $\pm$0.72} \\
        \bottomrule
    \end{tabular}
    \end{small}
    \vspace{-3mm}
\end{table*}

\begin{table*}[!htbp]
\caption{5-shot classification accuracies ($\%$) of different labeled ratios on \textit{mini}ImageNet.} 
\label{app:tab:classification_main_ratio_5_shot}
    \small
    \vspace{-3mm}
    \centering
    \begin{small} 
    \begin{tabular}{l|ccccc}
        \toprule
        & \multicolumn{4}{c}{\textit{5-shot}} \\
        Methods & 1\% & 10\% & 20\% & 30\% \\
        \midrule
       
        Soft k-Means~\cite{ren2018meta}         &	38.20{\tiny $\pm$1.64} &	56.27{\tiny $\pm$1.75} &	60.13{\tiny $\pm$1.79} &	62.47{\tiny $\pm$1.65}\\
        Soft k-Means+Cluster~\cite{ren2018meta} &	33.65{\tiny $\pm$1.53} &	56.87{\tiny $\pm$1.77} &	60.33{\tiny $\pm$1.81} &	62.43{\tiny $\pm$1.79}\\
        Masked Soft k-Means~\cite{ren2018meta}  & 	39.33{\tiny $\pm$1.55} &	57.20{\tiny $\pm$1.64} &	62.50{\tiny $\pm$1.78} &	63.00{\tiny $\pm$1.77} \\
        TPN-semi~\cite{liu2018learning}             & 	46.27{\tiny $\pm$1.67} &	60.55{\tiny $\pm$0.72} &	62.43{\tiny $\pm$0.69} &	63.10{\tiny $\pm$0.69}\\
        MAML~\cite{finn2017model}                 & 	60.22{\tiny $\pm$0.83} &	61.06{\tiny $\pm$0.81} &	63.18{\tiny $\pm$0.76} &	65.60{\tiny $\pm$0.82}   \\
        PL~\cite{lee2013pseudo}                   & 	60.64{\tiny $\pm$0.92} &	61.17{\tiny $\pm$0.85} &	63.06{\tiny $\pm$0.79} &	65.14{\tiny $\pm$0.74} \\
        \hline 
        \textsc{Gcmi} (ours)      &   63.62{\tiny $\pm$0.95} &	\textbf{62.72}{\tiny $\pm$0.88} &	64.78{\tiny $\pm$0.76} &	65.96{\tiny $\pm$0.74} \\
        \textsc{Flmi} (ours)      & \textbf{63.80}{\tiny $\pm$0.92} &	62.60{\tiny $\pm$0.86} &	\textbf{65.16}{\tiny $\pm$0.74} &	\textbf{66.10}{\tiny $\pm$0.79} \\
        \bottomrule
    \end{tabular}
    \end{small}
    \vspace{-3mm}
\end{table*}

\begin{figure}[ht]
\vskip 0.2in
\begin{center}
\centerline{\includegraphics[width=0.6\columnwidth]{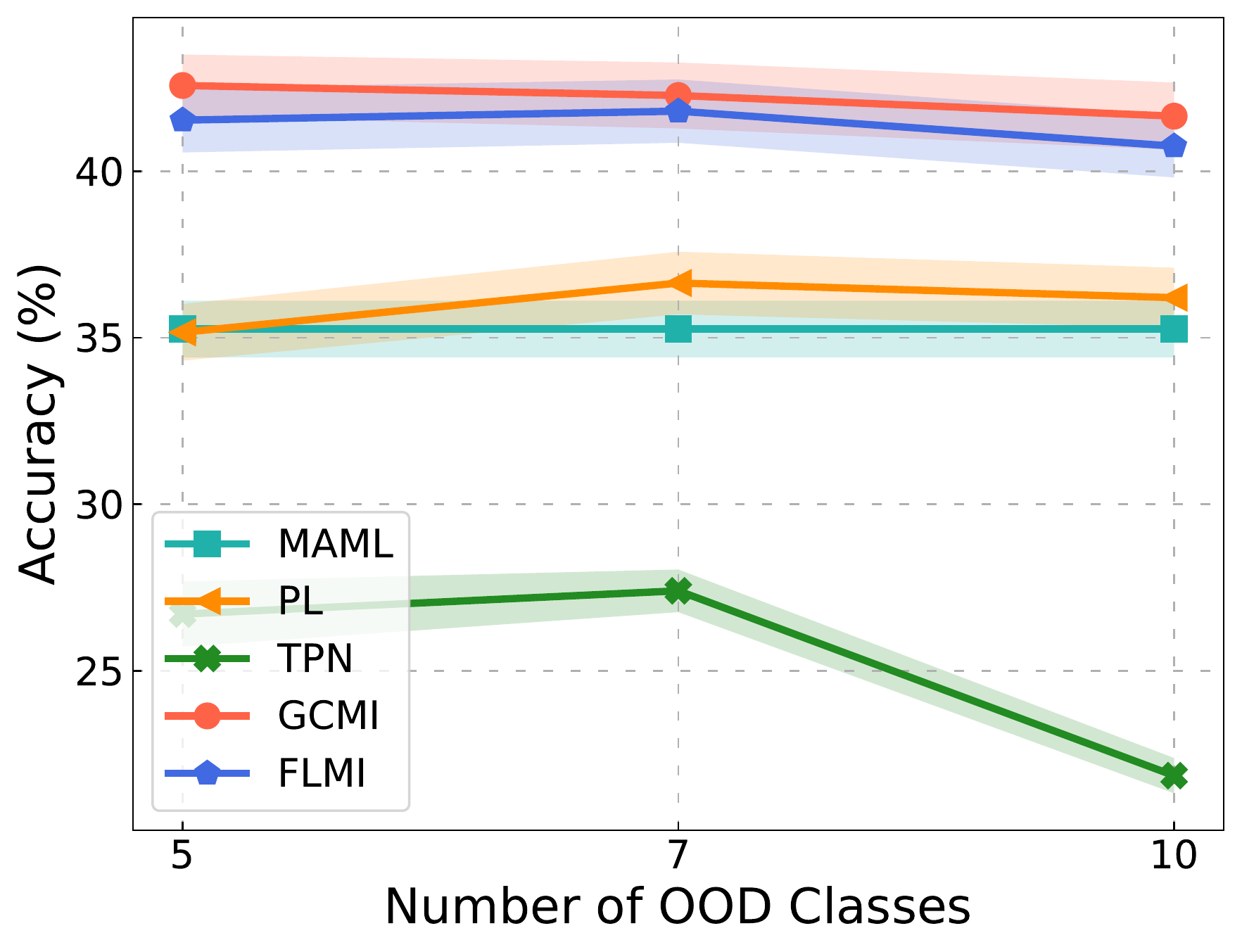}}
\caption{Comparison under different number of OOD classes in the unlabeled set for 1-shot case on \textit{mini}ImageNet. }
\label{fig:num_OOD_classes_1shot}
\end{center}
\vskip -0.2in
\end{figure}

\begin{figure}[ht]
\vskip 0.2in
\begin{center}
\centerline{\includegraphics[width=0.6\columnwidth]{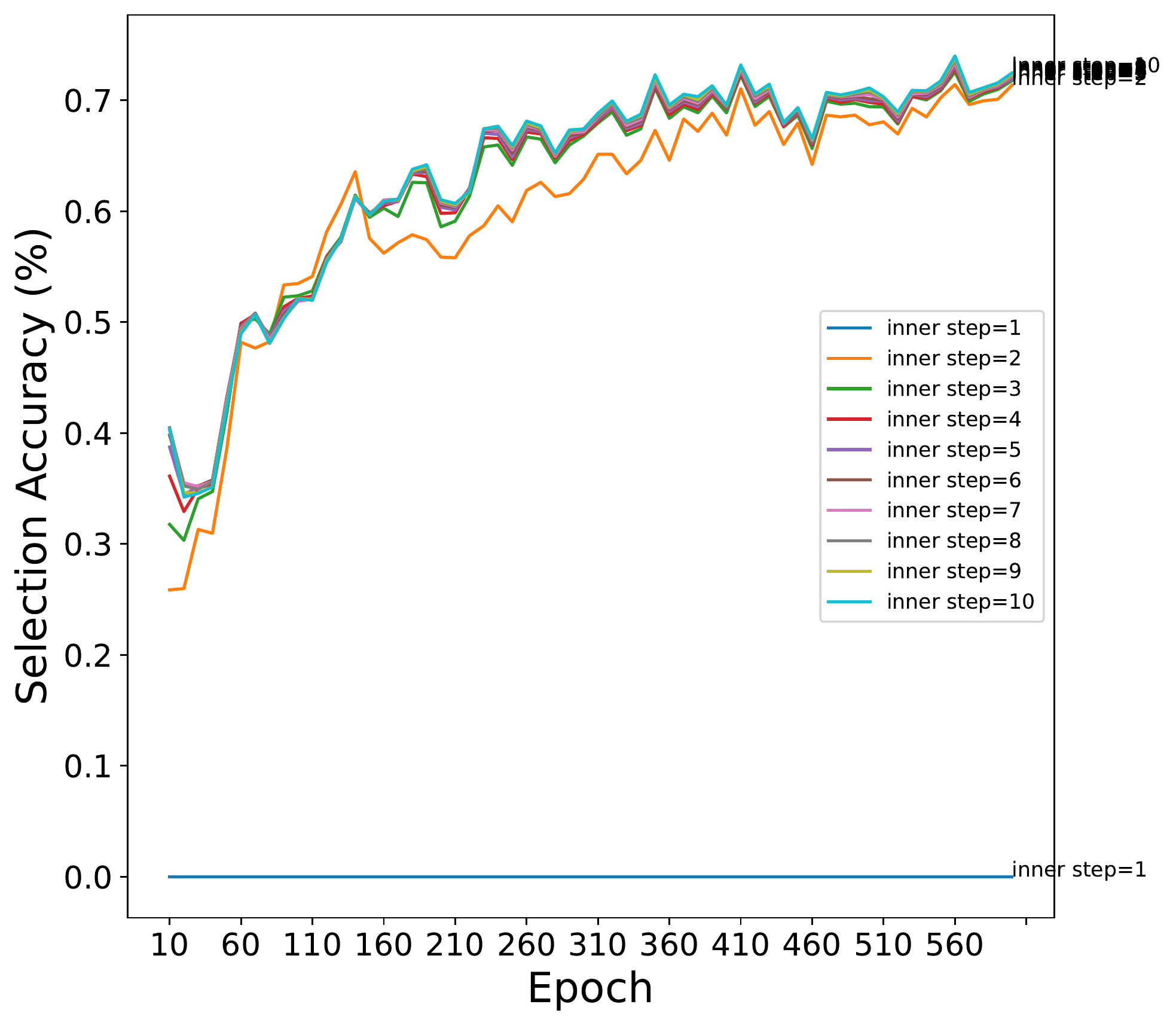}}
\caption{Selection accuracy in the unlabeled set for 1-shot case on \textit{mini}ImageNet during meta-test for PL. }
\label{fig:select_pl}
\end{center}
\vskip -0.2in
\end{figure}

\begin{figure}[ht]
\vskip 0.2in
\begin{center}
\centerline{\includegraphics[width=0.6\columnwidth]{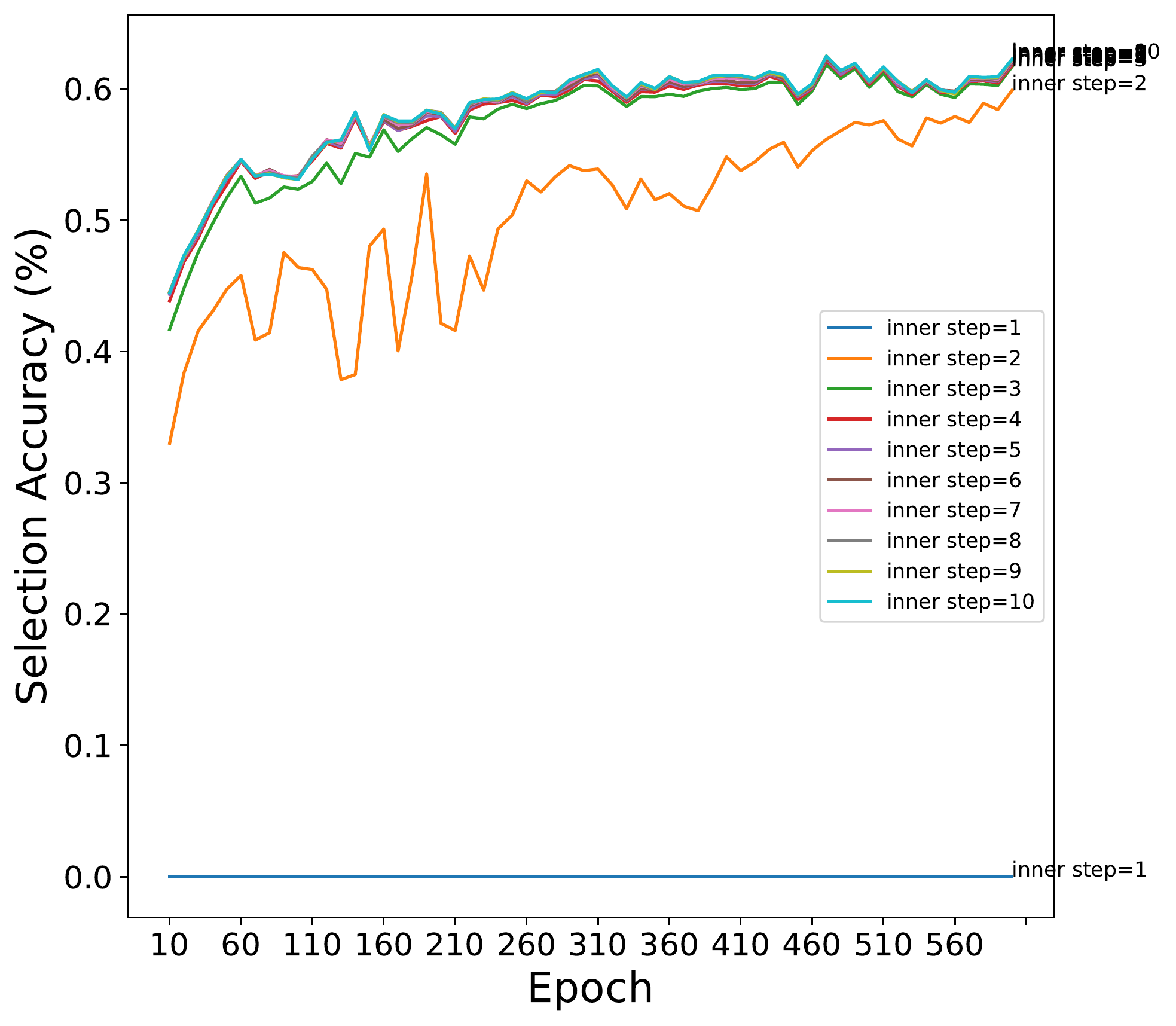}}
\caption{Selection accuracy in the unlabeled set for 1-shot case on \textit{mini}ImageNet during meta-test for \textsc{Gcmi}. }
\label{fig:select_gcmi}
\end{center}
\vskip -0.2in
\end{figure}

\begin{figure}[ht]
\vskip 0.2in
\begin{center}
\centerline{\includegraphics[width=0.6\columnwidth]{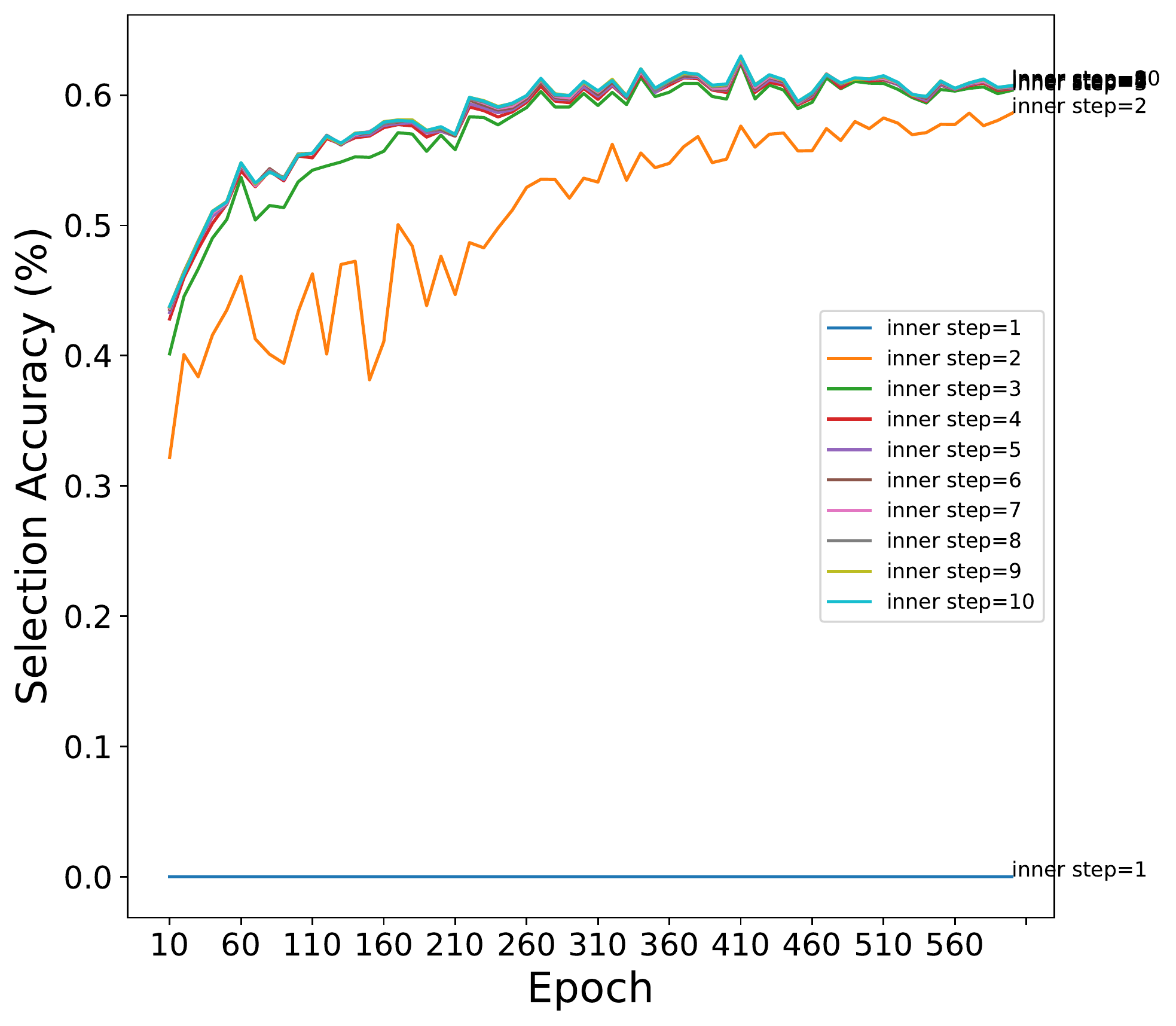}}
\caption{Selection accuracy in the unlabeled set for 1-shot case on \textit{mini}ImageNet during meta-test for \textsc{Flmi}. }
\label{fig:select_flmi}
\end{center}
\vskip -0.2in
\end{figure}

\textbf{Selection accuracy.}
\figref{fig:select_pl}, \figref{fig:select_gcmi}, and \figref{fig:select_flmi} show the selection accuracy of three SSL algorithms: PL, \textsc{Gcmi} and \textsc{Flmi} in the inner loop during meta-test. Although \textsc{Gcmi} and \textsc{Flmi} has slightly low accuracy than PL, this verifies that our proposed \model{} is able to select more balanced and diverse data which are more important for model training. 



\end{document}